\documentclass{article}

\PassOptionsToPackage{numbers, compress}{natbib}

\usepackage[final]{neurips_2023}

\usepackage[utf8]{inputenc} 
\usepackage[T1]{fontenc}    
\usepackage{url}            
\usepackage{booktabs}       
\usepackage{amsfonts}       
\usepackage{nicefrac}       
\usepackage{microtype}      
\usepackage{xcolor}         
\usepackage{xspace}
\usepackage{scale}

\usepackage{graphicx}
\usepackage{subfigure}
\usepackage{multirow}
\usepackage{svg}
\usepackage{float} 
\usepackage{longtable}
\usepackage{wrapfig}
\usepackage{caption}
\usepackage[ruled,noend,noline]{algorithm2e}
\usepackage{etoolbox}
\usepackage{amsmath}
\usepackage{newtxmath}

\usepackage[hidelinks]{hyperref}
\hypersetup{
    colorlinks=true,
    linkcolor=orange,
    filecolor=magenta,      
    urlcolor=orange,
    citecolor=orange,
}
\usepackage{pdfx}
\usepackage[capitalise, nameinlink]{cleveref}

\DeclareCaptionLabelFormat{table}{Table #2}
\makeatletter
\patchcmd{\@algocf@start}{-1.5em}{0em}{}{} 
\makeatother

\newcommand{\E}[0]{\mathbb{E}}
\newcommand{\stdv}[1]{\scalebox{0.7}{\!$\pm$#1}}
\newtheorem{theorem}{Theorem}
\newtheorem{lemma}{Lemma}
\newtheorem{proposition}{Proposition}
\newcommand{\logistic}[0]{\textrm{logistic}}

\newcommand{\fullname}[0]{Inverse Preference Learning\xspace}
\newcommand{\abv}[0]{IPL\xspace}

\title{\fullname:\\ Preference-based RL without a Reward Function}

\author{%
	Joey Hejna \\
        Stanford University \\
	\texttt{jhejna@cs.stanford.edu} \\
        \And
        Dorsa Sadigh \\
        Stanford University \\
        \texttt{dorsa@cs.stanford.edu} 
}

\begin{document}

\maketitle

\begin{abstract}
    Reward functions are difficult to design and often hard to align with human intent. Preference-based Reinforcement Learning (RL) algorithms address these problems by learning reward functions from human feedback. However, the majority of preference-based RL methods na\"ively combine supervised reward models with off-the-shelf RL algorithms. Contemporary approaches have sought to improve performance and query complexity by using larger and more complex reward architectures such as transformers. Instead of using highly complex architectures, we develop a new and parameter-efficient algorithm, \fullname (\abv), specifically designed for learning from offline preference data. Our key insight is that for a fixed policy, the $Q$-function encodes all information about the reward function, effectively making them interchangeable. Using this insight, we completely eliminate the need for a learned reward function. Our resulting algorithm is simpler and more parameter-efficient. Across a suite of continuous control and robotics benchmarks, \abv attains competitive performance compared to more complex approaches that leverage transformer-based and non-Markovian reward functions while having fewer algorithmic hyperparameters and learned network parameters. Our code is publicly released\footnote{\url{https://github.com/jhejna/inverse-preference-learning}}.
\end{abstract}
\section{Introduction}

Reinforcement Learning (RL) has shown marked success in fixed and narrow domains such as simulated control \citep{sac} and game-playing \citep{mnih2013playing}. When deploying RL in more complex settings, like in robotics or interaction with humans, one often runs into a critical bottleneck: the reward function. Obtaining reward labels in the real world can be complex, requiring difficult instrumentation \citep{pouring, zhu2019ingredients} and painstaking tuning \citep{yu2020meta} to achieve reasonable levels of sample efficiency. Moreover, despite extensive engineering, reward functions can still be exploited by algorithms in ways that do not align with human values and intents \citep{hadfield2017inverse}, which can be detrimental in safety-critical applications~\citep{amodei2016concrete}. 

Instead of hand-designing reward functions, contemporary works have attempted to learn them through expert demonstrations \citep{abbeel2004irl}, natural language \citep{lin2022inferring}, or human feedback \citep{sadigh2017active, akrour2011preference, wilson2012bayesian}. Recently, reward functions learned through pairwise comparison queries---where a user is asked which of two demonstrated behaviors they prefer---have been shown to be effective in both control \citep{preference_drl, sadigh2017active, pebble} and natural language domains \citep{stiennon2020learning}. This is often referred to as \emph{Reinforcement Learning with Human Feedback (RLHF)}. Reward functions learned via RLHF can directly capture human intent, while avoiding alternative and more expensive forms of human feedback such as expert demonstrations. Preference-based RL algorithms for RLHF often interleave reward-learning from comparisons with off-the-shelf RL algorithms.

While preference-based RL methods discover reward functions that are aligned with human preferences, they are not without flaws. Learned reward functions must have adequate coverage of both the state and action space to attain good downstream performance. Consequently, learning the reward function can be expensive, usually requiring thousands of labeled preference queries. To mitigate these challenges, recent works have proposed improving learned reward functions by adding inductive biases before optimization with RL. \citet{hejna2022fewshot} pretrain reward functions with meta-learning. \citet{park2022surf} use data augmentation. \citet{early2022non} and \citet{kim2023preference} make the reward function non-Markovian using recurrent or large transformer sequence model architectures respectively. Such approaches increase the upfront cost of preference-based RL by using additional data or compute. Moreover, these techniques still combine reward optimization with vanilla RL algorithms. Ultimately, this just adds an extra learned component to already notoriously delicate RL algorithms, further increasing hyper-parameter tuning overhead. Preference-based RL approaches often end up training up to four distinct neural networks independently: a critic (with up to two networks), an actor, and a reward function. This can be problematic as prediction errors cascade from the reward function, to the critic, and ultimately the actor causing high variance in downstream performance. To address these issues, we propose a parameter-efficient algorithm specifically designed for preference-based RL that completely eliminates the need to explicitly learn a reward function. In doing so, we reduce both complexity and compute cost.

The key insight of our work is that, under a fixed policy, the $Q$-function learned by off-policy RL algorithms captures the same information as the learned reward function. For example, both the $Q$-function and reward function encode information about how desirable a state-action pair is. This begs the question: why do we need to learn a reward function in the first place? Our proposed solution, \fullname or \abv, is an offline RL algorithm that is specifically designed for learning from preference data. Instead of relying on an explicit reward function, \abv directly optimizes the implicit rewards induced by the learned $Q$-function to be consistent with expert preferences. At the same time, \abv regularizes these implicit rewards to ensure high-quality behavior. As a result, \abv removes the need for a learned reward function and its associated computational and tuning expense. 

Experimentally, we find that even though \abv does not explicitly learn a reward function, it achieves competitive performance with complicated Transformer-based reward learning techniques on offline Preference-based RL benchmarks with real-human feedback. At the same time, \abv consistently exhibits lower variance across runs as it does not suffer from the errors associated with querying a learned reward model. Finally, under a minimal parameter budget, \abv is able to outperform standard preference-based RL approaches that learn an explicit reward model.

\section{Related Work}
Our work builds upon literature in reward learning, preference-based RL, and imitation learning.

\textbf{Reward Learning.} Due to the challenges associated with designing and shaping effective reward signals, several works have investigated various approaches for learning reward functions.
 A large body of work uses inverse RL to learn a reward function from expert demonstrations \citep{abbeel2004irl, ng2000irl, ramachandran2007bayesian}, which are unfortunately difficult to collect \citep{khurshid2015data, akgun2012keyframe, losey2020controlling} or often misaligned with true human preferences \citep{basu2017you,kwon2020humans}. Subsequently, reward learning techniques using other simpler forms of feedback such as scalar scores \citep{knox2008tamer} and partial \citep{myers2022learning} or complete rankings \citep{brown2019extrapolating,biyik2019green} have been developed. One of the simplest forms of human feedback is pairwise comparisons, where the user chooses between two options. Often, pairwise comparison queries are sampled using techniques from active learning  \citep{sadigh2017active,biyik2020active, daniel2015active}. However, to evaluate learned reward functions, these methods rely on either RL or traditional planning algorithms which are complex and computationally expensive. Our approach takes a simpler perspective that is parameter-efficient by combining reward and policy learning. Though it is not the focus of our work, \abv could additionally leverage active learning techniques for selecting preference data online.

\textbf{Preference-based Deep Reinforcement Learning.} Current approaches to preference based deep RL train a reward function, and then use that reward function in conjunction with a standard reinforcement learning algorithm \citep{preference_drl, lee2021bpref, shin2021offline}. Several techniques have been developed to improve the learned reward function, such as pre-training \citep{ibarz2018preference_demo, pebble}, meta-learning \citep{hejna2022fewshot}, data augmentation \citep{park2022surf}, and non-Markovian modeling. Within the family of non-Markovian reward modeling \citep{bacchus1996rewarding}, recent approaches have leveraged both LSTM networks \citep{early2022non} and transformers \citep{kim2023preference} for reward learning. But, these methods still rely on Markovian offline RL algorithms such as Implicit Q-Learning (IQL) \citep{kostrikov2021offline} for optimization. Ultimately, this makes such approaches theoretically inconsistent as the policy learning component assumes the reward to be only a function of the current state and action. All techniques for learning the reward function in combination with standard RL methods \citep{sac, ppo} end up adding additional hyper-parameter tuning and compute cost. \abv on the other hand, is directly designed for RL from preference data and eliminates the reward network entirely. Other recent works also consider contrastive objectives instead of RL \citep{pmlr-v202-kang23b, hejna2023contrastive}.
 
 Recently, works in natural language processing have applied ideas from preference-based RL to tasks such as summarization \citep{stiennon2020learning, wu2021recursively}, instruction following \citep{ouyang2022training}, and question-answering \citep{nakano2021webgpt}. The RLHF paradigm has proven to be powerful even at the massive scale of aligning large language models. In this regime, learned reward models are massive, making an implicit reward method like \abv more attractive. In fact, \abv in a contextual bandits setting recovers concurrent work by \citet{rafailov2023direct} on implicit reward modeling in LLMs (see \cref{app:theory}). While we focus on control in our experiments, we hope our work can inform future explorations in language domains.
	
\textbf{Imitation Learning}. Our work builds on foundational knowledge in maximum entropy (MaxEnt) RL \citep{ziebart2010modeling} and inverse RL \citep{ziebart2008maximum}. Recent works in MaxEnt inverse RL have used the mapping between $Q$-functions and reward functions under a fixed policy. Specifically, \citet{iqlearn} show that the regularized MaxEnt inverse RL objective from \citet{ho2016generative} can be re-written using the $Q$-function instead of a reward function and \citet{al-hafez2023lsiq} stabilize their approach. While the relationship between $Q$-functions and rewards has been used for MaxEnt inverse RL, we study this relationship when learning from preference data. While both problems seek to learn models of expert reward, the data differs significantly --- preference-based RL uses comparisons instead of optimal demonstrations. This necessitates a greatly different approach.

\section{Inverse Preference Learning}
In this section, we first describe the preference-based RL problem. Then, we describe how, leveraging techniques from imitation learning, we can remove the independently learned reward network from prior methods. This results in a simpler algorithm with lower computational cost and variance in performance. \looseness=-1
 
\subsection{Preference-Based RL}
We consider the reinfrocement leraning (RL) paradigm where an agent seeks to maximize its expected cumulative discounted sum of rewards in a Markov Decision Process (MDP). Standard off-policy RL algorithms, do so using state, action, reward, and next state tuples $(s,a,r,s')$. In preference-based RL, however, the reward function $r$ is unknown, and must be learned from human feedback. Thus, Traditional preference-based RL methods are thus usually separated into two stages: first, reward learning, where $r_E$ is estimated by a learned reward function $r_\theta$, and second, reinforcement learning, where a policy $\pi(a|s)$ is learned to maximize $\E_\pi[\sum_{t=0}^\infty \gamma^t r_\theta(s,a)]$ with $\gamma$ as the discount factor. Though our method combines these two phases, we use the building blocks of each and consequently review them here. 


\textbf{Preference Learning.} First, similar to prior works \citep{preference_drl, pebble}, we assume access to preference data in the form of binary comparisons. Each comparison is comprised of two behavior segments, $\sigma^{(1)}$ and $\sigma^{(2)}$, and a binary label $y$ indicating which of the two was preferred by an expert. As in \citet{wilson2012bayesian}, each behavior segment is simply a snippet of a trajectory of length $k$, or $\sigma = (s_t, a_t, s_{t+1}, a_{t+1}, \dots , a_{t+k-1}, s_{t+k} )$. Increasing $k$ can provide more information per label at the cost of potentially noisier labels. The label $y$ for each comparison is assumed to be generated by an expert according to a Bradley-Terry Preference model \citep{bradley1952rank}:
\begin{align}
\label{eq:bradley_terry}
P_{r_E}[\sigma^{(1)} \succ \sigma^{(2)}] = \frac{\exp \sum_t r_E(s_t^{(1)}, a_t^{(1)})}{\exp \sum_t r_E(s_t^{(1)}, a_t^{(1)}) + \exp \sum_t r_E(s_t^{(2)}, a_t^{(2)})},
\end{align}
where $r_E(s_t, a_t)$ is again the expert's unknown underlying reward model. We use the subscript $r_E$ on probability $P$ to indicate that the preference distribution above results from the expert's reward function. Let the dataset of these preferences be $\mathcal{D}_p = \{(\sigma^{(1)}, \sigma^{(2)}, y)\}$. To learn $r_E$, prior works in preference-based RL estimate a parametric reward function $r_\theta$ by minimizing the binary-cross-entropy over $\mathcal{D}_p$:
\begin{equation}
\label{eq:pref_loss}
\mathcal{L}_p(\theta) = -\E_{\sigma^{(1)}, \sigma^{(2)},y \sim \mathcal{D}_p}\left[y \log P_{r_\theta}\left[\sigma^{(1)} \succ \sigma^{(2)}\right] + (1-y) \log \left(1-P_{r_\theta}\left[\sigma^{(1)} \succ \sigma^{(2)}\right]\right)\right].
\end{equation}
This objective results from simply minimizing $\E_{\mathcal{D}_p}[D_{\text{KL}}(P_{r_E}||P_\theta)]$, the KL-divergence between the expert preference model and the one induced by $r_\theta$, effectively aligning it with the expert's preferences. We note that some other works in preference-based RL focus on learning an improved model $r_\theta$ to address the reward learning part of the problem \citep{park2022surf, kim2023preference}. However, these methods still use off-the-shelf RL algorithms for the policy learning part of the problem. 

\textbf{Reinforcement Learning.} Common off-policy RL methods learn a policy $\pi$ by alternating between policy evaluation (using the contractive Bellam Operator $\mathcal{B}_r^\pi$) to estimate $Q^\pi$ and policy improvement, where the policy $\pi$ is improved \citep{sutton2018reinforcement}. Concretely, after repeated application of $\mathcal{B}^\pi_r$ as
\begin{equation}
    \label{eq:bellman}
    (\mathcal{B}^\pi_r Q)(s,a) = r(s,a) + \gamma \E_{s' \sim p(\cdot|s,a)}[V^\pi(s')],
\end{equation}
the policy can be improved by maximizing $Q$. In some settings, the Bellman operator $\mathcal{B}^*_r$ corresponding to the optimal policy $\pi^*$ can be used directly, removing the need for the policy improvement step. In these cases, we can simply extract $\pi^*$ from the resulting $Q^*$.

To learn the optimal policy, two-phase preference based RL methods rely on recovering the optimal $r_E$ in the reward learning phase before running RL.  This potentially propagates errors from the estimated $r_\theta$ to the learned $Q$-function and ultimately learned policy $\pi$. In practice, it would be more efficient to eliminate the need for two separate stages. In the next section, we show how this can be done by establishing a bijection between reward functions $r$ and $Q$-functions. 

\begin{figure}[t]
\includegraphics[width=\textwidth]{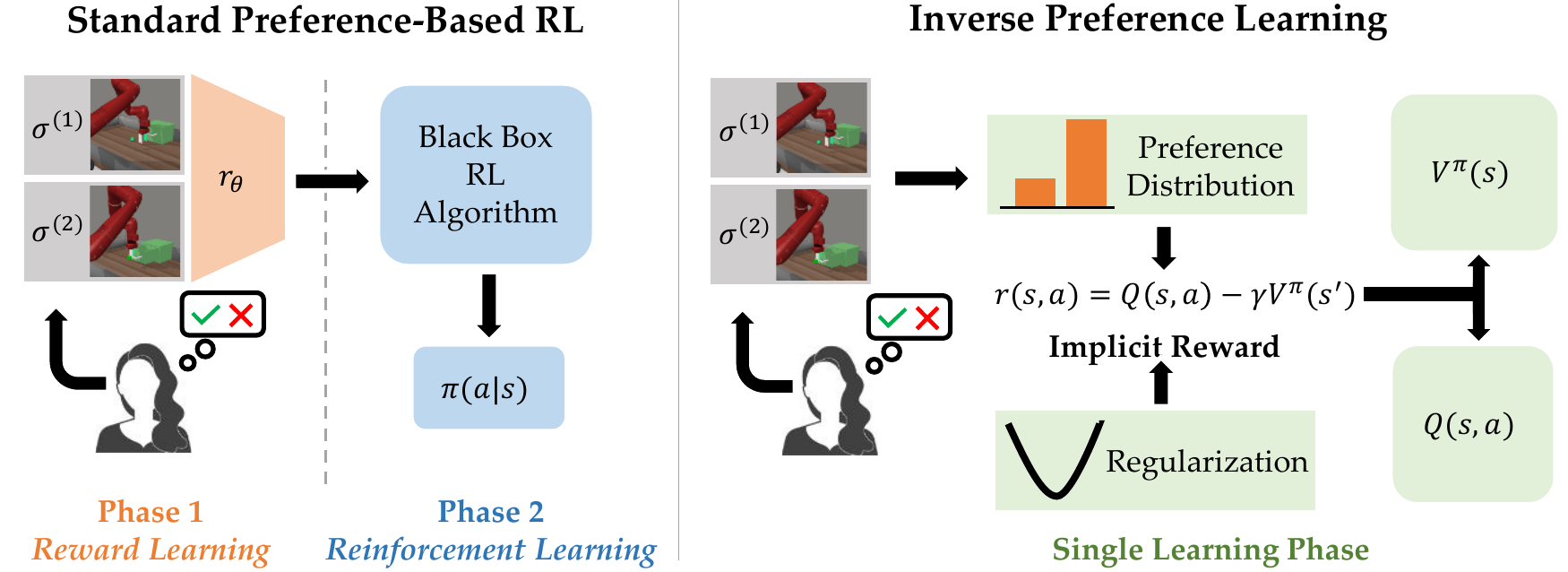}
\caption{A depiction of the difference between standard preference-based RL methods and \fullname. Standard preference-based RL first learns a reward function, then optimizes it with a blockbox RL algorithm. \abv trains a $Q$ function to directly fit the expert's preferences. This is done by aligning the implied reward model with the expert's preference distribution and applying regularization.}
\end{figure}

\subsection{Removing The Reward Function}
\label{sec:remove_reward}

In this section, we formally describe how the reward function can be removed from offline preference-based RL algorithms. Our key insight is that the $Q$-function learned by off-policy RL algorithms in fact encodes the same information as the reward function $r(s, a)$. Consequently, it is unnecessary to learn both. First, we show how the reward function can be re-written in terms of the $Q$ function allowing us to compute the preference model $P_Q$ induced by the $Q$-function. Then, we derive an objective that simultaneously pushes $Q$ to fit the expert's preferences while also remaining optimal. \looseness=-1

Consider fitting a $Q$ function via the Bellman operator $\mathcal{B}^\pi_r$ for a fixed policy $\pi$ until convergence where $\mathcal{B}^\pi_r Q = Q $. Here, to encode the cumulative discounted rewards when acting according to the policy, the $Q$-function depends on both $r$ and $\pi$. This dependence, however, is directly disentangled by the Bellman equation. By rearranging it (\cref{eq:bellman}), we can solve for the reward function in terms of $Q$ and $\pi$. This yields the so-called inverse soft-Bellman operator:
\begin{equation}
\label{eq:inverse_bellman}
(\mathcal{T}^\pi Q)(s,a) = Q(s,a) - \gamma \E_{s'}[V^\pi(s')].
\end{equation}
In fact, for a fixed policy $\pi$ the inverse-Bellman operator is bijective, implying a one-to-one correspondence between the $Q$ function and the reward function. Though this was previously shown in maximum entropy RL \citep{iqlearn}, we prove the general case \cref{lemma:bijection} in \cref{app:theory}.

Intuitively, this makes sense: when holding the policy constant, only the reward function affects $Q$. We abbreviate the evaluation of $(\mathcal{T}^\pi Q)(s,a)$ as $r_{Q^\pi}(s,a)$ to indicate that $r_{Q^\pi}$ is the unique implicit reward function induced by $Q^\pi$. Prior works in imitation learning leverage the inverse soft-Bellman operator to measure how closely the implicit reward model $r_{Q^\pi}$ aligns with expert demonstrations \citep{iqlearn}. Our key insight is that this equivalence can also be used to directly measure how closely our $Q$ function aligns with the expert preference model \textit{without ever directly learning $r$}. 

Consider the Bradley-Terry preference model in Equation \eqref{eq:bradley_terry}. For a fixed policy $\pi$ and its corresponding $Q^\pi$, we can obtain the preference model of the implicit reward function $P_{Q^\pi}[\sigma^{(1)} \succ \sigma^{(2)}]$ by substituting the inverse Bellman operator into Equation \eqref{eq:bradley_terry} as follows:
\begin{equation}
\label{eq:q_preference}
P_{Q^\pi}[\sigma^{(1)} > \sigma^{(2)}] = \frac{\exp \sum_t  (\mathcal{T}^\pi Q)(s_t^{(1)}, a_t^{(1)}) }{\exp \sum_t (\mathcal{T}^\pi Q)(s_t^{(1)}, a_t^{(1)}) + \exp \sum_t (\mathcal{T}^\pi Q)(s_t^{(2)}, a_t^{(2)})}.
\end{equation}
This substitution will allow us to measure the difference between the preferences implied by $Q^\pi$ and those of the expert. To minimize the difference, we can propagate gradients through the preference modeling loss (Equation \eqref{eq:pref_loss}) and the implicit preference model $P_{Q^\pi}$ (Equation~\eqref{eq:q_preference}) to $Q$---just as we would for a parameterized reward estimate $r_\theta$. Unfortunately, na\"ively performing this substitution is insufficient to solve the RL objective for two reasons.

\textbf{The Optimal Inverse Bellman Operator.}
First, we have used an arbitrary policy $\pi$, not the optimal one, for converting from $Q$-values to rewards. Though the $Q$-function may imply the expert's preferences, the corresponding policy could be extremely sub-optimal. To fix this problem, we need to use the optimal inverse bellman operator $\mathcal{T}^*$ to ensure the extract $Q$-function corresponds to that of $\pi^*$. For this step, we can use any off-policy RL-algorithm that converges to the optimal policy! If the algorithm directly estimates the $\mathcal{B}^*_r$, the corresponding $\mathcal{T}^*$ can be estimated using the target from $\mathcal{B}^*_r$, or
\begin{equation*}
    (\mathcal{T}^* Q)(s,a) = Q(s,a) - \gamma \E_{s'}[V^{\text{targ}}(s')] \text{ where } V^\text{targ}(s) \text{ is estimated as in } \mathcal{B}^*_r.
\end{equation*}
In many cases, however, computing the optimal bellman operator $\mathcal{B}^*_r$ is infeasible. Instead, many modern off-policy RL algorithms use policy improvement to converge to the optimal policy. These methods, like \citet{sac} use $Q^\pi$ to estimate a new policy $\pi'$ such that $Q^{\pi'} \geq Q^\pi$. By repeatedly improving the policy, they eventually converge to $Q^*$. Thus, by repeatedly improving the policy according to these algorithms, we can eventually converge to the optimal policy and can thus estimate corresponding optimal inverse bellman operator by using $V^\text{targ}(s) = \E_{a \sim \pi(\cdot|s)}[Q(s,a)]$ in the above equation. 

\textbf{Regularization.} Given we can estimate $\mathcal{T}^*$ using targets from $\mathcal{B}^*_r$ or policy improvement, we can fit the optimal $Q$-function by minimizing the following loss function
\begin{equation*}
\mathcal{L}_p(Q) = -\E_{\sigma^{(1)}, \sigma^{(2)},y \sim \mathcal{D}_p}\left[y \log P_{Q^*}[\sigma^{(1)} \succ \sigma^{(2)}] + (1-y) \log (1-P_{Q^*}[\sigma^{(1)} > \sigma^{(2)})\right].
\end{equation*}
where $P_{Q^*}$ is given by substituting $\mathcal{T}^*Q$ into \cref{eq:q_preference}. Unfortunately, optimizing this objective alone leads to poor results and may not converge when using RL algorithms that depend on policy improvement. This is because the above objective is under constrained due to the invariance of the Bradley-Terry preference model to shifts. By examining \cref{eq:bradley_terry}, it can be seen that adding a constant value to all rewards does not change the probability of preferring a segment. However, shifting the reward function by a constant \emph{does} change the $Q$-function. RL algorithms using policy improvement monotonically increase the $Q$-function until reaching the maximum at $Q^*$. Thus as the implicit reward continues to increase, $Q^*$ will continue to increase and may never be reached. To resolve this issue, we insure that the optima of the preference loss is unique by introducing a convex regularizer $\psi(\cdot)$ on the implicit rewards $r_{Q^\pi} = \mathcal{T^\pi}Q$, giving us the regularized preference loss:
\begin{equation}
\label{eq:ipl}
\resizebox{.942\textwidth}{!}{
$\mathcal{L}_p(Q) = -\E_{\sigma^{(1)}, \sigma^{(2)},y \sim \mathcal{D}_p}\left[y \log P_{Q^*}[\sigma^{(1)} \succ \sigma^{(2)}] + (1-y) \log (1-P_{Q^*}[\sigma^{(1)} > \sigma^{(2)})\right] + \lambda \psi\left( \mathcal{T}^*Q \right)$
}
\end{equation}
In practice we choose $\psi$ to be a form of L2 regularization as is commonly done in imitation learning \cite{iqlearn, al-hafez2023lsiq} to prevent unbounded reward values. $\lambda > 0$ is a hyperparameter that controls the strength of regularization. Besides allowing us to guarantee convergence, regularization has a number of benefits. It can help center the implicit reward near zero, which has been shown to beneficial for RL \citep{Engstrom2020Implementation}. Moreover, it encourages more realistic implicit rewards. For example, a reward function might change rapidly by large values when only small perturbations are applied to the state or action. Though such reward functions might be unrealistic, they are completely valid solutions of the inverse-Bellman operator. Adding regularization can help penalize large deviations in reward unless they drastically reduce the preference loss. Thus, the first term of \cref{eq:ipl} encourages the $Q$-function to match the expert's preferences, while the second term smooths the implied reward function and makes it unique.

Our final algorithm, which we call \fullname (\abv) fits the optimal policy corresponding to the regularized expert reward function by repeatedly minimizing $\mathcal{L}_p(Q)$ (\cref{eq:ipl}) and improving the value target used $V^{\text{targ}}$ with the update step from any off-policy RL algorithm. In this manner, \abv performs dynamic programming through the inverse bellman operator until convergence. In \cref{app:theory}, we prove the following Theorem.

\begin{theorem}
    Given an off-policy RL algorithm that convergences to the optimal policy $\pi^*_r$ for some reward function $r$ and regularizer $\psi$ such that \cref{eq:pref_loss} is strictly convex, \abv converges to $\pi^*_{r^*}$ corresponding to reward function $r^* = \arg \min_r \E_{\mathcal{D}_p}[D_{\text{KL}}(P_{r_E}||P_\theta)] + \lambda \psi(r)$.
\end{theorem}

The proof of the theorem essentially relies on the fact that for a fixed policy $\pi$, we can optimize $\mathcal{L}_p(Q)$ (\cref{eq:ipl}) to fit $r^*$. Then, we can update the policy (or target values $V^\text{targ}$) and optimize $\mathcal{L}_p(Q)$ again. Because $r^*$ is unique, we fit $r^*$ again the second time, but the $Q$-function has improved. There are many choices of regularizers where this holds. In tabular settings if $\psi(r) =r^2$, $\mathcal{L}_p(Q)$ reduces to L2 regularized logistic regression, which is strictly convex, guaranteeing convergence (\cref{app:theory}).

Effectively, \abv removes the need to learn a reward network, while still converging to similar solution as other preference-based RL algorithms. Learning a reward network requires more parameters and a completely separate optimization loop, increasing compute requirements. Moreover, an explicit reward model introduces a whole new suite of hyper-parameters that need to be tuned including the model architecture, capacity, learning rate, batch size, and stopping criterion. In fact, because human preference data is so difficult to collect, many approaches opt to use simple accuracy thresholds instead of validation criteria to decide when to stop training $r_\theta$ \citep{pebble}. All of these components make preference-based RL unreliable and high-variance. On the other hand, our method completely removes all of these parameters in exchange for a single $\lambda$ hyper-parameter that controls the regularization strength. Though we have theoretically derived \abv, in the next section we provide practical recipes for applying it to offline preference-based RL.

\subsection{\abv for Offline Preference-based RL}

\begin{wrapfigure}[12]{R}{0.5\textwidth}
\vspace{-0.2in}
\begin{algorithm}[H]
\caption{IPL Algorithm (XQL Variant)}
\label{alg:ipl}
\SetKwInOut{Input}{Input}
\Input{$\mathcal{D}_p$, $\mathcal{D}_o$, $\lambda$, $\alpha$}
\For{$i = 1, 2, 3, ...$}{
Sample batches $B_p \sim \mathcal{D}_p, B_o \sim \mathcal{D}_o$ \\
Update $Q$: $\min_Q \E_{B_p}[\mathcal{L}_p(Q)]$ (\cref{eq:ipl}) \\
Update $V$: $\min_V \E_{B_p \cup B_o}[e^{z}- z - 1]$ \\ 
\quad \quad \quad \quad \quad where $z = Q(s,a) - V(s))/\alpha$ \\
}
Finally, extract $\pi(a|s)$ via: \\ 
\quad \quad $\max_\pi \E_{\mathcal{D}_p \cup \mathcal{D}_o}[e^{(Q(s,a) - V(s))/\alpha} \log \pi(a|s)]$

\end{algorithm}
\end{wrapfigure}

In offline preference-based RL, we assume access to a fixed offline dataset $\mathcal{D}_o = \{(s,a,s')\}$ of interactions without reward labels generated by a reference policy $\mu(a|s)$ in addition to the preference dataset $\mathcal{D}_p$. Common approaches to offline RL seek to learn \textit{conservative} policies that do not stray too far away from the distribution of data generated by $\mu(a|s)$. This is critical to prevent the policy $\pi$ from reaching out-of-distribution states during deployment which can be detrimental to performance. In this section, we detail a practical version of \abv that uses the $\mathcal{X}$QL offline RL algorithm \citep{xql}. $\mathcal{X}$QL fits the KL-constrained RL objective 
\begin{equation*}
    \max_\pi \E_\pi \left[\sum_{t=t'}^\infty \gamma^t \left( r(s_t, a_t) - \alpha \log \frac{\pi(a_t|s_t)}{\mu(a_t|s_t)}\right)\right]
\end{equation*}
where $\alpha$ controls the magnitude of the KL-divergence penalty. The $\mathcal{X}$QL algorithm directly fits the optimal $Q$-function using the optimal soft-Bellman operator \citep{xql, xu2023offline}
\begin{equation*}
    (\mathcal{B}^*_r Q)(s,a) = r(s,a) + \gamma \E_{s'}[V^{\text{targ}}(s')],\text{ where } V^\text{targ}(s) = \alpha \log \E_{a \sim \mu(\cdot | s)}\left[e^{Q(s, a) / \alpha}\right].
\end{equation*}
In practice, $V^\text{targ}$ is estimated using the linex loss function over the current $Q$-function. Thus, to fit the optimal $Q$-function, \abv with $\mathcal{X}$QL alternates between minimizing the preference loss $\mathcal{L}_p(Q)$ (\cref{eq:ipl}), and updating a learned value function $V$ until they converge to $Q^*$ and $V^*$. Note that we are not limited to using just $\mathcal{D}_p$. Though the preference modeling part of $\mathcal{L}_p(Q)$ can only be optimized with preference data $\mathcal{D}_p$, the value function can be updated with offline data as well. In the presence of additional offline data, we find that updating the value function using $\mathcal{D}_p \cup \mathcal{D}_o$ leads to better performance. We approximate L2 regularization with the regularizer $\psi(r) = \E_{\mathcal{D}_p \cup \mathcal{D}_o}[r(s,a)^2]$, which imposes an L2 penalty across the support of the data. While one might try to use weight decay to emulate L2-regularization, doing so is difficult in practice as $\mathcal{T}^*Q$ depends on both the $Q$ network and the target network. We find that weighting the regularization equally between $\mathcal{D}_p$ and $\mathcal{D}_o$ performs well. After $Q$ and $V$ have converged, we can extract the policy using the closed form relationship $\pi^*(a|s) \propto \mu(a|s) \exp{\left((Q^*(s,a) - V^*(s)) / \alpha \right)}$ for KL-constrained RL as in \citet{xql, peng2019advantage}. The full algorithm for \abv with $\mathcal{X}$QL can be found in \ref{alg:ipl}. 

Though we have shown how \abv can be instantiated with $\mathcal{X}$QL, it is fully with other offline RL algorithms. In fact, \abv can also be used with online RL algorithms like SAC \citep{sac}. Critically, this makes the \abv framework general, as it can remove the need for reward modeling in nearly any preference-based RL setting. This makes \abv simpler and more efficient. In the next section, we show that \abv can attain the same performance as strong offline preference-based RL baselines, without learning a reward network.

\section{Experiments}
In this section, we aim to answer the following questions: First, how does \abv compare to prior preference-based RL algorithms on standard benchmarks? Second, how does \abv perform in extremely data-limited settings? And finally, how efficient is \abv in comparison to two-phase preference-based RL methods?

\subsection{Setup}
As discussed in the previous section, though we use a KL-constrained objective for our theoretical derivation, in practice we can construct versions of \abv based on any offline RL algorithm. In our experiments we evaluate \abv with Implicit Q-Learning (IQL) \citep{kostrikov2021offline}, since it has been used in prior offline preference-based RL works. This allows us to directly compare \abv by isolating its implicit reward component  and using the same exact hyper-parameters as prior works. Using \abv with IQL  amounts to updating the value function according to the asymmetric expectile loss function instead of the linex loss function. Concretely, this can be done by replacing the value update in \cref{alg:ipl} with $\min_V \E_{B_p \cup B_o}\left[\left|\tau - \vmathbb{1}(Q(s,a)-V(s) < 0)\right| \left(Q(s,a) - V(s)\right)^2\right]$ where $\tau$ is the expectile.

Inspired by \citet{park2022surf}, we introduce data augmentations that sample sub-sections of behavior segments $\sigma$ during training. While such augmentations are inapplicable to non-Markovian reward models, we find that they boost performance for Markovian reward models while also reducing the total number of state-action pairs per batch of preference data. This is important as \abv needs data from both $\mathcal{D}_p$ and $\mathcal{D}_o$ to regularize the implicit reward function. Additional experiment details and hyper-parameters can be found in the Appendix.

\begin{table}[t]
\centering
\begingroup
\def\arraystretch{1.2}
\resizebox{\textwidth}{!}{%
\begin{tabular}{lcccccccc}
\multirow{2}{*}{Dataset} & \textbf{IQL}              & \textbf{MR}                 & \textbf{LSTM}             & \textbf{PT}        & \textbf{BREX}         & \textbf{MR}              & \textbf{IPL}             \\ 
                         & (Oracle)                  & (from \citep{kim2023preference})    & (from \citep{kim2023preference})    & (from \citep{kim2023preference}) & (reimpl.)       & (reimpl.)         & (Ours)                 \\  \hline
hop-m-r                  & 83.06 \stdv{15.8}        & 11.56 \stdv{30.3}          & 57.88 \stdv{40.6}        & \textbf{84.54 \stdv{4.1}}  & 62.0 \stdv{20.3}& 70.20 \stdv{35.0}   & 73.57 \stdv{6.7}        \\
hop-m-e                  & 73.55 \stdv{41.5}        & 57.75 \stdv{23.7}          & 38.63 \stdv{35.6}        & 68.96 \stdv{33.9}     & 85.1 \stdv{8.0} & \textbf{103.0 \stdv{5.6}}     & 74.52 \stdv{10.1}       \\
walk-m-r                 & 73.11 \stdv{8.1}         & 72.07 \stdv{2.0}           & \textbf{77.00 \stdv{3.0}}         & 71.27 \stdv{10.3}   & 10.3 \stdv{5.4} & 68.79 \stdv{5.6}      & 59.92 \stdv{5.1}        \\
walk-m-e                 & 107.8 \stdv{2.2}        & \textbf{108.3 \stdv{3.9}}          & \textbf{110.4 \stdv{0.9}}        & \textbf{110.1 \stdv{0.2}}    &99.62 \stdv{3.0}& \textbf{109.1 \stdv{1.3}}     & \textbf{108.51 \stdv{0.6}}       \\
lift-ph                  & 96.75 \stdv{1.8}         & 84.75 \stdv{6.2}           & 91.50 \stdv{5.4}         & 91.75 \stdv{5.9}    & 96.6 \stdv{3.0} & \textbf{98.84 \stdv{2.3}}      & \textbf{97.60 \stdv{2.9}}        \\
lift-mh                  & 86.75 \stdv{2.8}         & \textbf{91.00 \stdv{2.8}}           & \textbf{90.8 \stdv{5.8}}         & 86.75 \stdv{6.0}   & 60.4 \stdv{25.1} & \textbf{90.04 \stdv{4.5}}      & \textbf{87.20 \stdv{5.3}}        \\
can-ph                   & 74.50 \stdv{6.8}         & 68.00 \stdv{9.1}           & 62.00 \stdv{10.9}        & 69.67 \stdv{5.9}      & 63.0 \stdv{20.3} & \textbf{76.40 \stdv{3.7}}      & \textbf{74.8 \stdv{2.4}}         \\
can-mh                   & 56.25 \stdv{8.8}         & 47.50 \stdv{3.5}           & 30.50 \stdv{8.7}         & 50.50 \stdv{6.5}    & 30.4 \stdv{23.0}& 53.6 \stdv{7.9}       & \textbf{57.6 \stdv{5.0}}         \\ \hline
Avg Std                  & 10.95                     & 10.2                        & 13.87                     & 9.08            & 13.77 & 8.23                   & \textbf{4.8}
\end{tabular}%
}
\endgroup
\vspace{0.05in}
\caption{Average normalized scores of all baselines on human-preference benchmarks from \citet{kim2023preference}. For the D4RL locomotion tasks ``hop'' corresponds to hopper, ``m'' to medium (training the data generating agent to 1/3 expert performance), ``r'' to replay buffer data, and ``e'' to data from the end of training. For the Robomimic tasks lift and can, ``ph'' corresponds to proficient human data and ``mh'' to multi-human data of differing optimality. The first four columns are taken from \citet{kim2023preference}. ``reimpl.'' is our reimplementation of Markovian Reward with IQL. The ``Avg Std'' row shows the average standard deviation across all eight environments. We run five seeds and report the final performance at the end of training like \citet{kostrikov2021offline}. Bolded values are within 95\% of the top performing method. Note that standard devaition values in the table were rounded for space. On some tasks IPL achieves higher performance earlier in training, which is not reflected above (See \cref{app:results}). We find that \abv outperforms PT on many environments, and also performs similarly to our implementation of MR despite not training a reward function.}
\label{tab:benchmark}
\vspace{-0.15in}
\end{table}

\subsection{How does \abv perform on preference-based RL benchmarks?}
We compare \abv to other offline preference-based RL approaches on D4RL Gym Locomotion \citep{fu2020d4rl} and Robosuite robotics \citep{robomimic2021} datasets with real-human preference data from \citet{kim2023preference}. We compare IQL-based \abv, with the same hyper-parameters, to various baselines that learn a reward model $r_\theta$ before optimization with IQL. Markovian Reward or MR denotes using a standard Markovian MLP reward model, like those used in \citet{preference_drl} and \citet{pebble}. Note that this is also equivalent to T-REX \citep{brown2019extrapolating} for offline RL. Non-Markovian Reward or NMR denotes using the non-Markovian LSTM based reward model from \citet{early2022non}. Preference Transformer (PT) is a state-of-the-art approach that leverages a large transformer architecture to learn a non-Markovian reward and preference weighting function. B-REX uses bayesian optimization to fit a linear reward function from predefined features \citep{brown2020safe}, which in our case are random Gaussian projections of the states and actions. For fairness, we also compare against our own implementation of IQL with a Markovian Reward function that uses the same data augmentation as \abv.  

Our results are summarized in \cref{tab:benchmark}. Starting with the first column, we see that preference-based RL methods are able to match IQL with the  ground truth reward function in many cases. On, several tasks, however, the MR implementation from \citet{kim2023preference} fairs rather poorly. The non-Markovian methods, (NMR and PT) improve performance. It is worth noting that on many tasks our implementation of a MR (sixth column) performs far better than reported in \citet{kim2023preference}, likely due to our careful tuning of $r_\theta$ and use of data-augmentations. Our method, \abv, achieves competitive performance across the board. 

In general, \abv with IQL performs on-par or better than both our implementation of MR and PT in most datasets despite not learning a separate reward network. Specifically, \abv has the same performance or better performance than our MR implementation on six of eight tasks. More importantly, \abv does extremely well in comparison to Preference Transformer's reported results. On five of eight tasks \abv performs better than PT while having over 10 times fewer parameters, making \abv far more efficient. To be consistent with \citet{kim2023preference}, we report results after a million training steps but performance for \abv often peaks earlier (see learning curves in the Appendix). For example, with early stopping \abv also outperforms PT on ``hop-m-r''. We posit that this is because the $Q$-function in \abv is tasked with both fitting the expert's preference model and optimal policy simultaneously, making both the policy and reward function non-stationary during training. In some datasets, this was more unstable. 

\abv also has the lowest average standard-deviation across seeds, meaning it yields more consistent results than explicit reward methods. For standard two-phase preference-based RL algorithms, errors in the reward model are propagated to and exacerbated by the $Q$ function. \abv circumvents this problem by not explicitly learning the reward. 

Finally, in \cref{tab:ablations}, we consider various design decisions of \abv. Augmentations provide a strong boost in the robotics environment, but offer only minor improvements in locomotion. Removing regularization, however, is detrimental to performance. This is likely because without regularization, the implicit reward values can continue to increase, leading to exploding estimates of $Q$. Finally, we show that \abv is compatible with other offline RL algorithms by combining it with $\mathcal{X}$QL \citep{xql}. We find that with $\mathcal{X}$QL, \abv performs even better on some tasks, but worse on others. Finally, in \cref{app:results}, we also show that \abv can be combined with online preference-based RL algorithms like PEBBLE \citep{pebble}.

\begin{table}[t]
    \centering
    \begin{tabular}{lcccc}
    \textbf{Dataset}  & \textbf{No Aug}          & $\lambda = 0$    & \textbf{IPL-XQL }         & \textbf{IPL}              \\ \hline
    hop-m-r  & 70.46 \stdv{6.7} & 10.41 \stdv{2.26} & 80.4 \stdv{2.13}  & 73.57 \stdv{6.67} \\
    walk-m-r & 58.50 \stdv{5.3} & 4.85 \stdv{1.52}  & 57.82 \stdv{5.24} & 59.92 \stdv{5.11} \\
    lift-mh          & 84.8 \stdv{4.1}   & 52.60 \stdv{10.1}      & 89.00 \stdv{4.4}  & 87.20 \stdv{5.3}  \\
    can-mh   & 53.2 \stdv{5.8}  & 13.8 \stdv{5.7}  & 59.0 \stdv{5.0}   & 57.6 \stdv{5.00} 
    \end{tabular}
    \caption{Ablations for \abv on the offline human-preference benchmark. We consider removing data augmentation, removing regularization $\lambda = 0$, and other offline RL algorithms ($\mathcal{X}$QL). Full results can be found in \cref{app:results}.}
    \label{tab:ablations}
\end{table}

\subsection{How does \abv scale with Data?}
Collecting preference comparisons is often viewed as the most expensive part of preference-based RL. To investigate how well \abv performs in data limited settings, we construct scripted preference datasets of four different sizes for five tasks from the MetaWorld benchmark \citep{yu2020meta} used in prior preference-based RL works \citep{pebble,hejna2022fewshot}. We then train on the preference data $\mathcal{D}_p$ by setting $\mathcal{D}_o = \{(s,a,s') \in \mathcal{D}_p\}$ and use the same hyper-parameters for all environments and methods where applicable. Our results are summarized in~\cref{tab:data}. Again, \abv is a strong reward-free baseline. We find that at all data scales, \abv performs competitively to our implementation of MR (IQL with a learned Markovian reward) and consistently outperforms it in Button Press and Assembly.  Increasing the amount of preference data generally improves performance across the board. However, as we generate queries  uniformly at random some preference datasets may be easier to learn from than others, leading to deviations from this trend in some cases. As in the benchmark results in \cref{tab:benchmark}, \abv exhibits lower variance across seeds and tasks, in this case at three of four data scales. 

\begin{table}[t]
\centering
\begingroup
\def\arraystretch{1.05}
\begin{tabular}{llcccc}
\multicolumn{2}{r}{Preference Queries} & 500             & 1000             & 2000             & 4000           \\ [0.1cm] \hline 
\multirow{2}{*}{Button Press}   & MR   & \textbf{66.0 \stdv{8.0}} & 49.3 \stdv{12.1} & 54.7 \stdv{26.8} & 78.3 \stdv{9.2} \\
                                & IPL  & 53.3 \stdv{8.5} & \textbf{60.1 \stdv{12.8}} & \textbf{70.2 \stdv{2.5}}  & \textbf{90.2 \stdv{6.5}} \\ [0.1cm]
\multirow{2}{*}{Drawer Open}    & MR   & \textbf{65.9 \stdv{9.9}} & \textbf{87.2 \stdv{5.2}}  & \textbf{89.7 \stdv{6.4}}  & \textbf{94.6 \stdv{3.9}} \\
                                & IPL  & \textbf{62.1 \stdv{4.8}} & 78.7 \stdv{12.4} & \textbf{89.5 \stdv{5.0}}  & \textbf{96.6 \stdv{1.3}} \\ [0.1cm]
\multirow{2}{*}{Sweep Into}     & MR   & \textbf{33.0 \stdv{5.7}} & \textbf{46.2 \stdv{6.0}}  & \textbf{63.2 \stdv{13.7}} & \textbf{70.8 \stdv{7.9}} \\
                                & IPL  & \textbf{34.5 \stdv{2.3}} & \textbf{48.2 \stdv{7.2}}  & 58.8 \stdv{7.4}  & 65.9 \stdv{6.7} \\ [0.1cm]
\multirow{2}{*}{Plate Slide}    & MR   & \textbf{54.6 \stdv{5.3}} & \textbf{57.2 \stdv{4.5}}  & 23.9 \stdv{18.8} & \textbf{55.2 \stdv{3.0}} \\
                                & IPL  & \textbf{52.9 \stdv{4.8}} & \textbf{55.8 \stdv{2.2}}  & \textbf{55.4 \stdv{3.1}}  & \textbf{54.9 \stdv{2.8}} \\ [0.1cm]
\multirow{2}{*}{Assembly}       & MR   & 0.6 \stdv{0.7}  & 0.7 \stdv{1.0}   & 0.0 \stdv{0.0}   & 2.6 \stdv{2.8}  \\
                                & IPL  & \textbf{0.9 \stdv{0.6}}  & \textbf{1.5 \stdv{1.5}}   & \textbf{1.7 \stdv{1.9}}   & \textbf{5.5 \stdv{5.2}}  \\ [0.1cm] \hline 
\multirow{2}{*}{Avg Std}        & MR   & 5.9             & \textbf{5.76}             & 13.14            & 5.36           \\
                                & IPL  & \textbf{4.2}            & 7.22             & \textbf{3.98}             & \textbf{4.5}           
\end{tabular}
\endgroup
\caption{Results on five MetaWorld tasks at four different preference data scales. We run five seeds for each method, and take the highest average performance across seeds from the learning curves. More details can be found in \cref{app:results}. \abv performs the same or better than IQL with a Markovian reward model on the majority of tasks and preference data scales without training a reward model. \looseness=-1}
\label{tab:data}
\vspace{-0.15in}
\end{table}

\begin{figure}[t]
\centering
\includegraphics[width=\textwidth]{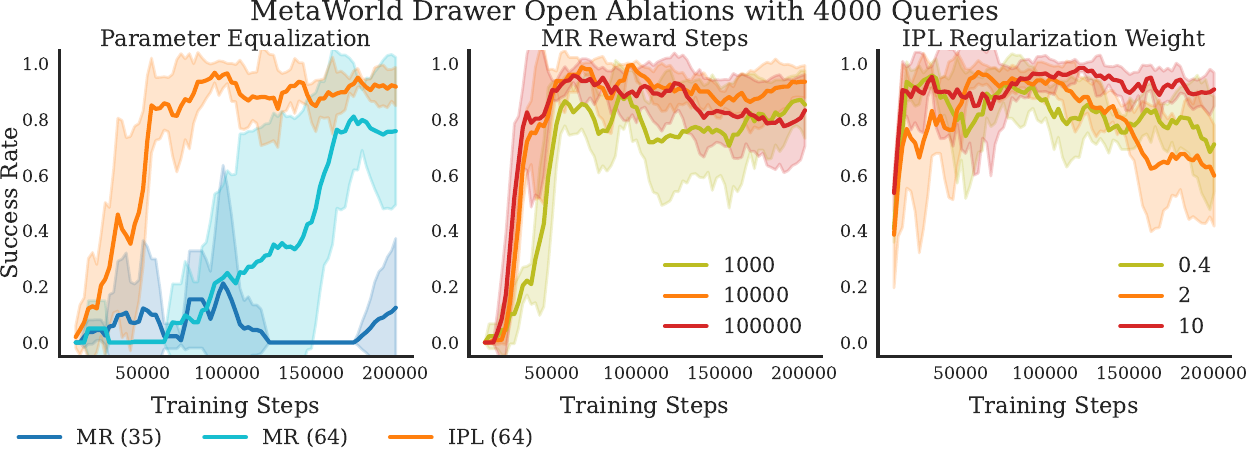}
\vspace{-0.1in}
\caption{\textbf{Left:} Performance comparison with different parameter numbers. MR (35) has the same parameter budget as \abv (64). MR (64) has over twice as many. We see that with the same number of parameters as \abv, MR is unable to adequetly fit the data and performs poorly. \textbf{Middle:} MR when the reward function is trained for a varying number of steps -- with too few the reward model under-fits, and with too many it over-fits, both leading to worse performance. \textbf{Right:} \abv with different regularization strengths. On the drawer open task, performance is largely unaffected. For more ablations, see the Appendix.}
\label{fig:ablation}

\end{figure}

\begin{wraptable}[16]{R}{0.3\textwidth}
\centering
\begin{tabular}{lr}
\textbf{Method}        & \textbf{Params}   \\ \hline
PT            & 2942218      \\
NMR           & 508746       \\
MR            & 348426       \\
IPL           & 278537       \\
MR  (64) & 34892             \\
IPL (64) & 14025        \\
MR  (35) & 14012                         
\end{tabular}
\vspace{-0.05in}
\caption{Parameter counts for different methods. The bottom three rows are for the limited parameter budget experiments in Section 4.4.}
\label{tab:params}
\end{wraptable}

\subsection{How efficient is \abv?}
One benefit of \abv over other preference-based RL methods  is its parameter efficiency. By removing the reward network, \abv uses fewer parameters than other methods while achieving the same performance. In \cref{tab:params}, we show the number of parameters for each method used in the last two sections. Preference Transformer uses over ten times more parameters than \abv, and the LSTM-based NMR model from \citet{early2022non} uses nearly twice as many. When dealing with a limited compute or memory budget, this can be important. To exacerbate this effect, we consider an extremely parameter efficient version of \abv, denoted ``IPL (64)'' in \cref{tab:params}, based on Advantage Weighted Actor Critic (AWAC) \citep{nair2021awac} which eliminates the second critic and value networks used in IQL \citep{kostrikov2021offline} and uses a two-layer 64-dimensional MLP. We then compare this parameter-efficient \abv to MR with the same parameter budget which results in ``MR (35)'', a 35-dimensional MLP. Results are depicted on the left of  \cref{fig:ablation}. MR trained with a smaller network is unable to adequately fit the data, resulting in lower performance. Only after increasing the network size past that of \abv can MR begin to match performance. \looseness=-1

Aside from parameter efficiency, \abv is also ``hyper-parameter efficient''. By removing the reward network, \abv removes a whole set of hyper-parameters associated with two phase preference based RL methods, like reward network architecture, learning rate, stopping criterion, and more. In the middle of \cref{fig:ablation} we show how the performance of MR is affected when the reward function is over or under fit. Choosing the correct number of steps to train the reward model usually requires collecting a validation set of preference data, which is costly to obtain. Instead of this, \abv only has a single regularization parameter, $\lambda$. The right side of \cref{fig:ablation} shows the sensitivity of \abv to $\lambda$. We find that in many cases, varying $\lambda$ has little effect on performance unless it is perturbed by a large amount.

\section{Conclusion}
\textbf{Summary.} We introduce \fullname, a novel algorithm for offline preference-based RL that avoids learning a reward function. Our key insight is to leverage the inverse soft-Bellman operator, which computes the mapping from $Q$-functions to rewards under a fixed policy. The \abv algorithm  trains a $Q$-function to regress towards the optimal $Q^*$ while at the same time admitting implicit reward values that are consistent with an expert's preferences. Even though \abv does not require learning a separate reward network, on robotics benchmarks it attains competitive performance with preference-based RL baselines that use twice to ten-times the number of model parameters. \looseness=-1 

\textbf{Limitations and Future Work.} A number of future directions remain. Specifically, the implicit reward function and policy learned by \abv are both non-stationary during training, which sometimes causes learning to be more unstable than with a fixed reward function. This is a core limitation future work could address by better mixing policy improvement and preference-matching steps to improve stability. More broadly, implicit reward preference-based RL methods are not limited to continuous control or binary feedback. Applying implicit reward techniques to other forms of feedback or extending \abv to language-based RLHF tasks remain exciting future directions. 


\newpage

\begin{ack}
    This work was supported by ONR, DARPA YFA, Ford, and NSF Awards \#1941722 and \#2218760. JH is supported by by the National Defense Science Engineering Graduate (NDSEG) Fellowship Program. We would additionally like to thank Div Garg and Chris Cundy for useful discussions.
\end{ack}

\bibliography{references}

\begin{thebibliography}{59}
\providecommand{\natexlab}[1]{#1}
\providecommand{\url}[1]{\texttt{#1}}
\expandafter\ifx\csname urlstyle\endcsname\relax
  \providecommand{\doi}[1]{doi: #1}\else
  \providecommand{\doi}{doi: \begingroup \urlstyle{rm}\Url}\fi

\bibitem[Abbeel and Ng(2004)]{abbeel2004irl}
Pieter Abbeel and Andrew~Y Ng.
\newblock Apprenticeship learning via inverse reinforcement learning.
\newblock In \emph{International Conference on Machine Learning}, 2004.

\bibitem[Akgun et~al.(2012)Akgun, Cakmak, Jiang, and Thomaz]{akgun2012keyframe}
Baris Akgun, Maya Cakmak, Karl Jiang, and Andrea~L Thomaz.
\newblock Keyframe-based learning from demonstration.
\newblock \emph{International Journal of Social Robotics}, 4\penalty0
  (4):\penalty0 343--355, 2012.

\bibitem[Akrour et~al.(2011)Akrour, Schoenauer, and
  Sebag]{akrour2011preference}
Riad Akrour, Marc Schoenauer, and Michele Sebag.
\newblock Preference-based policy learning.
\newblock In \emph{Joint European Conference on Machine Learning and Knowledge
  Discovery in Databases}, 2011.

\bibitem[Al-Hafez et~al.(2023)Al-Hafez, Tateo, Arenz, Zhao, and
  Peters]{al-hafez2023lsiq}
Firas Al-Hafez, Davide Tateo, Oleg Arenz, Guoping Zhao, and Jan Peters.
\newblock {LS}-{IQ}: Implicit reward regularization for inverse reinforcement
  learning.
\newblock In \emph{The Eleventh International Conference on Learning
  Representations}, 2023.
\newblock URL \url{https://openreview.net/forum?id=o3Q4m8jg4BR}.

\bibitem[Amodei et~al.(2016)Amodei, Olah, Steinhardt, Christiano, Schulman, and
  Man{\'e}]{amodei2016concrete}
Dario Amodei, Chris Olah, Jacob Steinhardt, Paul Christiano, John Schulman, and
  Dan Man{\'e}.
\newblock Concrete problems in ai safety.
\newblock \emph{arXiv preprint arXiv:1606.06565}, 2016.

\bibitem[Bacchus et~al.(1996)Bacchus, Boutilier, and
  Grove]{bacchus1996rewarding}
Fahiem Bacchus, Craig Boutilier, and Adam Grove.
\newblock Rewarding behaviors.
\newblock In \emph{National Conference on Artificial Intelligence}, 1996.

\bibitem[Basu et~al.(2017)Basu, Yang, Hungerman, Sinahal, and
  Draqan]{basu2017you}
Chandrayee Basu, Qian Yang, David Hungerman, Mukesh Sinahal, and Anca~D Draqan.
\newblock Do you want your autonomous car to drive like you?
\newblock In \emph{2017 12th ACM/IEEE International Conference on Human-Robot
  Interaction (HRI}, pages 417--425. IEEE, 2017.

\bibitem[B{\i}y{\i}k et~al.(2019)B{\i}y{\i}k, Lazar, Sadigh, and
  Pedarsani]{biyik2019green}
Erdem B{\i}y{\i}k, Daniel~A Lazar, Dorsa Sadigh, and Ramtin Pedarsani.
\newblock The green choice: Learning and influencing human decisions on shared
  roads.
\newblock In \emph{2019 IEEE 58th conference on decision and control (CDC)},
  pages 347--354. IEEE, 2019.

\bibitem[Biyik et~al.(2020)Biyik, Huynh, Kochenderfer, and
  Sadigh]{biyik2020active}
Erdem Biyik, Nicolas Huynh, Mykel~J. Kochenderfer, and Dorsa Sadigh.
\newblock Active preference-based gaussian process regression for reward
  learning.
\newblock In \emph{Proceedings of Robotics: Science and Systems (RSS)}, July
  2020.

\bibitem[Bradley and Terry(1952)]{bradley1952rank}
Ralph~Allan Bradley and Milton~E Terry.
\newblock Rank analysis of incomplete block designs: I. the method of paired
  comparisons.
\newblock \emph{Biometrika}, 39\penalty0 (3/4):\penalty0 324--345, 1952.

\bibitem[Brown et~al.(2019)Brown, Goo, Nagarajan, and
  Niekum]{brown2019extrapolating}
Daniel Brown, Wonjoon Goo, Prabhat Nagarajan, and Scott Niekum.
\newblock Extrapolating beyond suboptimal demonstrations via inverse
  reinforcement learning from observations.
\newblock In \emph{International conference on machine learning}, pages
  783--792. PMLR, 2019.

\bibitem[Brown et~al.(2020)Brown, Coleman, Srinivasan, and
  Niekum]{brown2020safe}
Daniel Brown, Russell Coleman, Ravi Srinivasan, and Scott Niekum.
\newblock Safe imitation learning via fast bayesian reward inference from
  preferences.
\newblock In \emph{International Conference on Machine Learning}, pages
  1165--1177. PMLR, 2020.

\bibitem[Christiano et~al.(2017)Christiano, Leike, Brown, Martic, Legg, and
  Amodei]{preference_drl}
Paul~F Christiano, Jan Leike, Tom Brown, Miljan Martic, Shane Legg, and Dario
  Amodei.
\newblock Deep reinforcement learning from human preferences.
\newblock In \emph{Advances in Neural Information Processing Systems}, 2017.

\bibitem[Daniel et~al.(2015)Daniel, Kroemer, Viering, Metz, and
  Peters]{daniel2015active}
Christian Daniel, Oliver Kroemer, Malte Viering, Jan Metz, and Jan Peters.
\newblock Active reward learning with a novel acquisition function.
\newblock \emph{Autonomous Robots}, 39\penalty0 (3):\penalty0 389--405, 2015.

\bibitem[Early et~al.(2022)Early, Bewley, Evers, and Ramchurn]{early2022non}
Joseph Early, Tom Bewley, Christine Evers, and Sarvapali Ramchurn.
\newblock Non-markovian reward modelling from trajectory labels via
  interpretable multiple instance learning.
\newblock In \emph{Advances in Neural Information Processing Systems}, 2022.

\bibitem[Engstrom et~al.(2020)Engstrom, Ilyas, Santurkar, Tsipras, Janoos,
  Rudolph, and Madry]{Engstrom2020Implementation}
Logan Engstrom, Andrew Ilyas, Shibani Santurkar, Dimitris Tsipras, Firdaus
  Janoos, Larry Rudolph, and Aleksander Madry.
\newblock Implementation matters in deep rl: A case study on ppo and trpo.
\newblock In \emph{International Conference on Learning Representations}, 2020.
\newblock URL \url{https://openreview.net/forum?id=r1etN1rtPB}.

\bibitem[Fu et~al.(2020)Fu, Kumar, Nachum, Tucker, and Levine]{fu2020d4rl}
Justin Fu, Aviral Kumar, Ofir Nachum, George Tucker, and Sergey Levine.
\newblock D4rl: Datasets for deep data-driven reinforcement learning.
\newblock \emph{arXiv preprint arXiv:2004.07219}, 2020.

\bibitem[Garg et~al.(2021)Garg, Chakraborty, Cundy, Song, and Ermon]{iqlearn}
Divyansh Garg, Shuvam Chakraborty, Chris Cundy, Jiaming Song, and Stefano
  Ermon.
\newblock Iq-learn: Inverse soft-q learning for imitation.
\newblock In \emph{Thirty-Fifth Conference on Neural Information Processing
  Systems}, 2021.
\newblock URL \url{https://openreview.net/forum?id=Aeo-xqtb5p}.

\bibitem[Garg et~al.(2023)Garg, Hejna, Geist, and Ermon]{xql}
Divyansh Garg, Joey Hejna, Matthieu Geist, and Stefano Ermon.
\newblock Extreme q-learning: Maxent {RL} without entropy.
\newblock In \emph{The Eleventh International Conference on Learning
  Representations}, 2023.
\newblock URL \url{https://openreview.net/forum?id=SJ0Lde3tRL}.

\bibitem[Haarnoja et~al.(2018)Haarnoja, Zhou, Abbeel, and Levine]{sac}
Tuomas Haarnoja, Aurick Zhou, Pieter Abbeel, and Sergey Levine.
\newblock Soft actor-critic: Off-policy maximum entropy deep reinforcement
  learning with a stochastic actor.
\newblock In \emph{International Conference on Machine Learning}, 2018.

\bibitem[Hadfield-Menell et~al.(2017)Hadfield-Menell, Milli, Abbeel, Russell,
  and Dragan]{hadfield2017inverse}
Dylan Hadfield-Menell, Smitha Milli, Pieter Abbeel, Stuart~J Russell, and Anca
  Dragan.
\newblock Inverse reward design.
\newblock \emph{Advances in neural information processing systems}, 30, 2017.

\bibitem[Hejna and Sadigh(2022)]{hejna2022fewshot}
Joey Hejna and Dorsa Sadigh.
\newblock Few-shot preference learning for human-in-the-loop {RL}.
\newblock In \emph{Conference on Robot Learning}, 2022.

\bibitem[Hejna et~al.(2023)Hejna, Rafailov, Sikchi, Finn, Niekum, Knox, and
  Sadigh]{hejna2023contrastive}
Joey Hejna, Rafael Rafailov, Harshit Sikchi, Chelsea Finn, Scott Niekum,
  W~Bradley Knox, and Dorsa Sadigh.
\newblock Contrastive preference learning: Learning from human feedback without
  rl.
\newblock \emph{arXiv preprint arXiv:2310.13639}, 2023.

\bibitem[Ho and Ermon(2016)]{ho2016generative}
Jonathan Ho and Stefano Ermon.
\newblock Generative adversarial imitation learning.
\newblock \emph{Advances in neural information processing systems}, 29, 2016.

\bibitem[Ibarz et~al.(2018)Ibarz, Leike, Pohlen, Irving, Legg, and
  Amodei]{ibarz2018preference_demo}
Borja Ibarz, Jan Leike, Tobias Pohlen, Geoffrey Irving, Shane Legg, and Dario
  Amodei.
\newblock Reward learning from human preferences and demonstrations in atari.
\newblock In \emph{Advances in Neural Information Processing Systems}, 2018.

\bibitem[Kang et~al.(2023)Kang, Shi, Liu, He, and Wang]{pmlr-v202-kang23b}
Yachen Kang, Diyuan Shi, Jinxin Liu, Li~He, and Donglin Wang.
\newblock Beyond reward: Offline preference-guided policy optimization.
\newblock In Andreas Krause, Emma Brunskill, Kyunghyun Cho, Barbara Engelhardt,
  Sivan Sabato, and Jonathan Scarlett, editors, \emph{Proceedings of the 40th
  International Conference on Machine Learning}, volume 202 of
  \emph{Proceedings of Machine Learning Research}, pages 15753--15768. PMLR,
  23--29 Jul 2023.
\newblock URL \url{https://proceedings.mlr.press/v202/kang23b.html}.

\bibitem[Khurshid and Kuchenbecker(2015)]{khurshid2015data}
Rebecca~P Khurshid and Katherine~J Kuchenbecker.
\newblock Data-driven motion mappings improve transparency in teleoperation.
\newblock \emph{Presence}, 24\penalty0 (2):\penalty0 132--154, 2015.

\bibitem[Kim et~al.(2023)Kim, Park, Shin, Lee, Abbeel, and
  Lee]{kim2023preference}
Changyeon Kim, Jongjin Park, Jinwoo Shin, Honglak Lee, Pieter Abbeel, and Kimin
  Lee.
\newblock Preference transformer: Modeling human preferences using transformers
  for rl.
\newblock In \emph{International Conference on Learning Representations}, 2023.

\bibitem[Knox and Stone(2008)]{knox2008tamer}
W~Bradley Knox and Peter Stone.
\newblock Tamer: Training an agent manually via evaluative reinforcement.
\newblock In \emph{2008 7th IEEE international conference on development and
  learning}, pages 292--297. IEEE, 2008.

\bibitem[Kostrikov et~al.(2022)Kostrikov, Nair, and
  Levine]{kostrikov2021offline}
Ilya Kostrikov, Ashvin Nair, and Sergey Levine.
\newblock Offline reinforcement learning with implicit q-learning.
\newblock In \emph{International Conference on Learning Representations}, 2022.

\bibitem[Kwon et~al.(2020)Kwon, Biyik, Talati, Bhasin, Losey, and
  Sadigh]{kwon2020humans}
Minae Kwon, Erdem Biyik, Aditi Talati, Karan Bhasin, Dylan~P Losey, and Dorsa
  Sadigh.
\newblock When humans aren’t optimal: Robots that collaborate with risk-aware
  humans.
\newblock In \emph{2020 15th ACM/IEEE International Conference on Human-Robot
  Interaction (HRI)}, pages 43--52. IEEE, 2020.

\bibitem[Lee et~al.(2021{\natexlab{a}})Lee, Smith, and Abbeel]{pebble}
Kimin Lee, Laura Smith, and Pieter Abbeel.
\newblock Pebble: Feedback-efficient interactive reinforcement learning via
  relabeling experience and unsupervised pre-training.
\newblock In \emph{International Conference on Machine Learning},
  2021{\natexlab{a}}.

\bibitem[Lee et~al.(2021{\natexlab{b}})Lee, Smith, Dragan, and
  Abbeel]{lee2021bpref}
Kimin Lee, Laura Smith, Anca Dragan, and Pieter Abbeel.
\newblock B-pref: Benchmarking preference-based reinforcement learning.
\newblock In \emph{Conference on Neural Information Processing Systems Datasets
  and Benchmarks Track (round 1)}, 2021{\natexlab{b}}.

\bibitem[Lin et~al.(2022)Lin, Fried, Klein, and Dragan]{lin2022inferring}
Jessy Lin, Daniel Fried, Dan Klein, and Anca Dragan.
\newblock Inferring rewards from language in context.
\newblock \emph{arXiv preprint arXiv:2204.02515}, 2022.

\bibitem[Losey et~al.(2020)Losey, Srinivasan, Mandlekar, Garg, and
  Sadigh]{losey2020controlling}
Dylan~P Losey, Krishnan Srinivasan, Ajay Mandlekar, Animesh Garg, and Dorsa
  Sadigh.
\newblock Controlling assistive robots with learned latent actions.
\newblock In \emph{2020 IEEE International Conference on Robotics and
  Automation (ICRA)}, pages 378--384. IEEE, 2020.

\bibitem[Mandlekar et~al.(2021)Mandlekar, Xu, Wong, Nasiriany, Wang, Kulkarni,
  Fei-Fei, Savarese, Zhu, and Mart\'{i}n-Mart\'{i}n]{robomimic2021}
Ajay Mandlekar, Danfei Xu, Josiah Wong, Soroush Nasiriany, Chen Wang, Rohun
  Kulkarni, Li~Fei-Fei, Silvio Savarese, Yuke Zhu, and Roberto
  Mart\'{i}n-Mart\'{i}n.
\newblock What matters in learning from offline human demonstrations for robot
  manipulation.
\newblock In \emph{Conference on Robot Learning (CoRL)}, 2021.

\bibitem[Mnih et~al.(2013)Mnih, Kavukcuoglu, Silver, Graves, Antonoglou,
  Wierstra, and Riedmiller]{mnih2013playing}
Volodymyr Mnih, Koray Kavukcuoglu, David Silver, Alex Graves, Ioannis
  Antonoglou, Daan Wierstra, and Martin Riedmiller.
\newblock Playing atari with deep reinforcement learning.
\newblock \emph{arXiv preprint arXiv:1312.5602}, 2013.

\bibitem[Myers et~al.(2022)Myers, Biyik, Anari, and Sadigh]{myers2022learning}
Vivek Myers, Erdem Biyik, Nima Anari, and Dorsa Sadigh.
\newblock Learning multimodal rewards from rankings.
\newblock In \emph{Conference on Robot Learning}, pages 342--352. PMLR, 2022.

\bibitem[Nair et~al.(2021)Nair, Dalal, Gupta, and Levine]{nair2021awac}
Ashvin Nair, Murtaza Dalal, Abhishek Gupta, and Sergey Levine.
\newblock {\{}AWAC{\}}: Accelerating online reinforcement learning with offline
  datasets, 2021.
\newblock URL \url{https://openreview.net/forum?id=OJiM1R3jAtZ}.

\bibitem[Nakano et~al.(2021)Nakano, Hilton, Balaji, Wu, Ouyang, Kim, Hesse,
  Jain, Kosaraju, Saunders, et~al.]{nakano2021webgpt}
Reiichiro Nakano, Jacob Hilton, Suchir Balaji, Jeff Wu, Long Ouyang, Christina
  Kim, Christopher Hesse, Shantanu Jain, Vineet Kosaraju, William Saunders,
  et~al.
\newblock Webgpt: Browser-assisted question-answering with human feedback.
\newblock \emph{arXiv preprint arXiv:2112.09332}, 2021.

\bibitem[Ng et~al.(2000)Ng, Russell, et~al.]{ng2000irl}
Andrew~Y Ng, Stuart~J Russell, et~al.
\newblock Algorithms for inverse reinforcement learning.
\newblock In \emph{International Conference on Machine Learning}, 2000.

\bibitem[Ouyang et~al.(2022)Ouyang, Wu, Jiang, Almeida, Wainwright, Mishkin,
  Zhang, Agarwal, Slama, Ray, et~al.]{ouyang2022training}
Long Ouyang, Jeff Wu, Xu~Jiang, Diogo Almeida, Carroll~L Wainwright, Pamela
  Mishkin, Chong Zhang, Sandhini Agarwal, Katarina Slama, Alex Ray, et~al.
\newblock Training language models to follow instructions with human feedback.
\newblock \emph{arXiv preprint arXiv:2203.02155}, 2022.

\bibitem[Park et~al.(2022)Park, Seo, Shin, Lee, Abbeel, and Lee]{park2022surf}
Jongjin Park, Younggyo Seo, Jinwoo Shin, Honglak Lee, Pieter Abbeel, and Kimin
  Lee.
\newblock Surf: Semi-supervised reward learning with data augmentation for
  feedback-efficient preference-based reinforcement learning.
\newblock In \emph{International Conference on Learning Representations}, 2022.

\bibitem[Peng et~al.(2019)Peng, Kumar, Zhang, and Levine]{peng2019advantage}
Xue~Bin Peng, Aviral Kumar, Grace Zhang, and Sergey Levine.
\newblock Advantage-weighted regression: Simple and scalable off-policy
  reinforcement learning.
\newblock \emph{arXiv preprint arXiv:1910.00177}, 2019.

\bibitem[Rafailov et~al.(2023)Rafailov, Sharma, Mitchell, Ermon, Manning, and
  Finn]{rafailov2023direct}
Rafael Rafailov, Archit Sharma, Eric Mitchell, Stefano Ermon, Christopher~D
  Manning, and Chelsea Finn.
\newblock Direct preference optimization: Your language model is secretly a
  reward model.
\newblock \emph{arXiv preprint arXiv:2305.18290}, 2023.

\bibitem[Ramachandran and Amir(2007)]{ramachandran2007bayesian}
Deepak Ramachandran and Eyal Amir.
\newblock Bayesian inverse reinforcement learning.
\newblock In \emph{IJCAI}, volume~7, pages 2586--2591, 2007.

\bibitem[Sadigh et~al.(2017)Sadigh, Dragan, Sastry, and
  Seshia]{sadigh2017active}
Dorsa Sadigh, Anca~D Dragan, Shankar Sastry, and Sanjit~A Seshia.
\newblock Active preference-based learning of reward functions.
\newblock In \emph{Robotics: Science and Systems}, 2017.

\bibitem[Schenck and Fox(2017)]{pouring}
C.~Schenck and D.~Fox.
\newblock Visual closed-loop control for pouring liquids.
\newblock In \emph{International Conference on Robotics and Automation}, 2017.

\bibitem[Schulman et~al.(2017)Schulman, Wolski, Dhariwal, Radford, and
  Klimov]{ppo}
John Schulman, Filip Wolski, Prafulla Dhariwal, Alec Radford, and Oleg Klimov.
\newblock Proximal policy optimization algorithms.
\newblock \emph{arXiv preprint arXiv:1707.06347}, 2017.

\bibitem[Shin and Brown(2021)]{shin2021offline}
Daniel Shin and Daniel~S Brown.
\newblock Offline preference-based apprenticeship learning.
\newblock \emph{arXiv preprint arXiv:2107.09251}, 2021.

\bibitem[Stiennon et~al.(2020)Stiennon, Ouyang, Wu, Ziegler, Lowe, Voss,
  Radford, Amodei, and Christiano]{stiennon2020learning}
Nisan Stiennon, Long Ouyang, Jeff Wu, Daniel~M Ziegler, Ryan Lowe, Chelsea
  Voss, Alec Radford, Dario Amodei, and Paul Christiano.
\newblock Learning to summarize from human feedback.
\newblock \emph{arXiv preprint arXiv:2009.01325}, 2020.

\bibitem[Sutton and Barto(2018)]{sutton2018reinforcement}
Richard~S Sutton and Andrew~G Barto.
\newblock \emph{Reinforcement learning: An introduction}.
\newblock MIT Press, 2018.

\bibitem[Wilson et~al.(2012)Wilson, Fern, and Tadepalli]{wilson2012bayesian}
Aaron Wilson, Alan Fern, and Prasad Tadepalli.
\newblock A bayesian approach for policy learning from trajectory preference
  queries.
\newblock In \emph{Advances in Neural Information Processing Systems}, 2012.

\bibitem[Wu et~al.(2021)Wu, Ouyang, Ziegler, Stiennon, Lowe, Leike, and
  Christiano]{wu2021recursively}
Jeff Wu, Long Ouyang, Daniel~M Ziegler, Nissan Stiennon, Ryan Lowe, Jan Leike,
  and Paul Christiano.
\newblock Recursively summarizing books with human feedback.
\newblock \emph{arXiv preprint arXiv:2109.10862}, 2021.

\bibitem[Xu et~al.(2023)Xu, Jiang, Li, Yang, Wang, Chan, and
  Zhan]{xu2023offline}
Haoran Xu, Li~Jiang, Jianxiong Li, Zhuoran Yang, Zhaoran Wang, Victor Wai~Kin
  Chan, and Xianyuan Zhan.
\newblock Offline rl with no ood actions: In-sample learning via implicit value
  regularization.
\newblock In \emph{International Conference on Learning Representations}, 2023.

\bibitem[Yu et~al.(2020)Yu, Quillen, He, Julian, Hausman, Finn, and
  Levine]{yu2020meta}
Tianhe Yu, Deirdre Quillen, Zhanpeng He, Ryan Julian, Karol Hausman, Chelsea
  Finn, and Sergey Levine.
\newblock Meta-world: A benchmark and evaluation for multi-task and meta
  reinforcement learning.
\newblock In \emph{Conference on Robot Learning}, 2020.

\bibitem[Zhu et~al.(2020)Zhu, Yu, Gupta, Shah, Hartikainen, Singh, Kumar, and
  Levine]{zhu2019ingredients}
Henry Zhu, Justin Yu, Abhishek Gupta, Dhruv Shah, Kristian Hartikainen, Avi
  Singh, Vikash Kumar, and Sergey Levine.
\newblock The ingredients of real world robotic reinforcement learning.
\newblock In \emph{International Conference on Learning Representations}, 2020.

\bibitem[Ziebart(2010)]{ziebart2010modeling}
Brian~D Ziebart.
\newblock \emph{Modeling purposeful adaptive behavior with the principle of
  maximum causal entropy}.
\newblock Carnegie Mellon University, 2010.

\bibitem[Ziebart et~al.(2008)Ziebart, Maas, Bagnell, Dey,
  et~al.]{ziebart2008maximum}
Brian~D Ziebart, Andrew~L Maas, J~Andrew Bagnell, Anind~K Dey, et~al.
\newblock Maximum entropy inverse reinforcement learning.
\newblock In \emph{Aaai}, volume~8, pages 1433--1438. Chicago, IL, USA, 2008.

\end{thebibliography}
\bibliographystyle{plainnat}


\newpage
\appendix
\section*{Appendix}

\section{Theory}
\setcounter{theorem}{0} 

\label{app:theory}

\subsection{Proofs}
\begin{lemma}
    \label{lemma:bijection}
    For any fixed policy $\pi$ the inverse bellman operator $\mathcal{T}^\pi$ establishes a bijection between $r$ and $Q$. Moreover, for any $r$, $Q = (\mathcal{T}^\pi)^{-1}r$ is the unique fixed point of the Bellman operator $\mathcal{B}_r^\pi$. (Adapted from \citet{iqlearn})
\end{lemma}

\textit{Proof}. Let $P^\pi$ be the stochastic transition matrix for the MDP corresponding to a fixed policy $\pi$. In vector form, the inverse bellman operator becomes $r = \mathcal{T}^\pi Q = (I - \gamma P^\pi) Q$. We can establish a bijection between $Q$ and $r$ by showing that $(I - \gamma P^\pi)$ is invertible. As $P^\pi$ defines a valid probability distribution over next stat-action pairs and $\gamma < 1$, we have that $||\gamma P^\pi|| < 1$. Thus, its Neumann series convergences, which implies the existence of $(I - \gamma P^\pi)^{-1}$. So, $Q = (I - \gamma P^\pi)^{-1} r$ and a bijection exists. Using this, we can also show a 1-1 mapping with the bellman operator under reward $r$. We have $Q = (I - \gamma P^\pi)^{-1} r = (\mathcal{T}^\pi)^{-1}r = \mathcal{B}^\pi_r Q$ at the fixed point of $\mathcal{B}^\pi_r$. 

\begin{theorem}
    Given an off-policy RL algorithm that convergences to the optimal policy $\pi^*_r$ for some reward function $r$ and regularizer $\psi$ such that \cref{eq:pref_loss} is strictly convex, \abv converges to $\pi^*_{r^*}$ corresponding to reward function $r^* = \arg \min_r \E_{\mathcal{D}_p}[D_{\text{KL}}(P_{r_E}||P_\theta)] + \lambda \psi(r)$.
\end{theorem}

\textit{Proof.} We prove this statement in the Tabular setting, first for algorithms that use policy improvement. Let $Q_t \in \mathbb{R}^{|S \times A|}$ and $\pi_t$ indicate the Q-function and policy after $t$ iterations. Let $Q_0 = 1/(1 - \gamma) \min_{S \times A} r(s,a)$. The inverse bellman operator tells us, in vector form, that $r = (I - \gamma P^\pi)Q$  where $P^\pi$ is the transition matrix. Let $r^* = \arg \min_r \mathbb{E}_{D_p}[y \log P_r + (1-y) \log (1 - P_r)]  + \lambda \psi(r)$, or the minimizer of the preference loss with regularizer $\psi$ such that we converge to a unique $r^*$.

At each step of IPL, we substitute the inverse bellman operator into the preference loss and optimize. Thus at convergence, $(I - \gamma P^{\pi_t})Q_t = r^*$ uniquely due to Lemma 1. Then, there are two cases based on the type of RL algorithm.

If our RL algorithm can directly estimate $\mathcal{B}^*_r$, then we are done. This is because we have assumed convergence, and thus $\mathcal{T}^*Q = r^*$. By the bijection established in Lemma 1, we have that $Q = (\mathcal{T}^*)^{-1}r^*$ which by also Lemma 1 is the unique fixed point of $\mathcal{B}^*_{r^*}$ which is $Q^*_{r^*}$. Thus, we have recovered the optimal $Q$ function for $r^*$ from which the optimal policy $\pi^*_{r^*}$ can be extracted.

If we have an RL algorithm that uses policy improvement, we consider multiple steps of \abv. If the RL algorithm guarantees convergence via policy improvement, then we use $\pi_t$ and $Q_t$. to obtain a new policy $\pi_{t+1}$. Using $\pi_{t+1}$ we can obtain the transition matrix $P^{\pi_{t+1}}$. 
Finally, we optimize the preference loss again using $P^{\pi_{t+1}}$ in the inverse Bellman operator to obtain $Q_{t+1}$. At convergence $ (I - \gamma P^{\pi_{t+1}})Q_{t+1} = r^*$ holds. 
As $r^*$ is unique, $Q_t$ and $Q_{t+1}$ are both Q-functions for the reward function $r^*$, just under different policies. We know from the definition of policy improvement, that $Q^{\pi_{t+1}} \geq Q^{\pi_{t}}$ necessarily, and thus $Q_{t+1} \geq Q_{t}$ for any $t$. Convergence is possible as according to Lemma 1, $Q^* = (\mathcal{T}^*)^{-1} r$ is the a fixed point of $\mathcal{B}^*_r$.

\begin{proposition}
    If $\phi(r) = r^2$, then \abv converges to the optimal policy corresponding to  $r^* = \arg \min_r \E_{\mathcal{D}_p}[D_{\text{KL}}(P_{r_E}||P_\theta)] + \lambda \psi(r)$ in tabular settings.
\end{proposition}

\textit{Proof.} The preference-based loss function with L2 regularization can be viewed as L2 regularized logistic regression by writing the logits as a dot-product between a preference comparison vector $x$ comprised of $-\gamma^t, \gamma^t$ and 0 terms and a reward function $r$. The Hessian for this objective is then $X^T D X + \lambda I \quad \text{ where } D_{ii} = \logistic(x_i \top r) (1 - \logistic(x_i \top r))$, which is positive definite. Thus the problem is strictly convex and $r^*$ is unique, so \abv converges to its optimal policy by Theorem 1. Note that to guarantee this we must regularize $r$ across the entire state-action space, analogous to regularizing the weight vector in logistic regression.

\subsection{Connections to DPO}
Concurrent work, called Direct Preference Optimization (DPO) \citep{rafailov2023direct} also remove the need for explicit reward modeling for learning from preferences, but do so in the contextual bandits setting for language models. Thus, DPO is limited to preference queries segment length 1 that \textit{must start from the same state}. \abv is in fact, a more general version of DPO that does not have these restrictions. Specifically, \abv with $\mathcal{X}$QL recovers the same exact policy as DPO when applied to the contextual bandits setting. 

Within the bandits setting, there is no  “next-state” and $V^*(s')$ is removed, and the inverse bellman operator becomes just $Q(s,a) = r(s,a)$. The optimal $\mathcal{X}$QL policy is $\pi^* = \mu(a|s) e^{Q^*(s,a)}/ Z(s)$ where $Z$ is the partition function. By rearranging, $\mathcal{T}^*Q = Q^*(s,a) = \log \frac{\pi(a|s)}{\mu(a|s)} + Z(s)$. We can plug this into the preference model induced by Q in \cref{eq:q_preference}. In the RLHF setting, the partition function cancels since we assume the context to be the same between preferences. This exactly results in the DPO algorithm, showing the DPO is in fact just an instantiation of \abv for contextual bandits.

\subsection{\abv with Rankings}

\abv can easily be extended to rankings using a Plackett Luce model. Consider permutations $\tau$ over $K$ segments: 

\begin{equation*}
    P_{r_E}(\tau) = \prod_{k=1}^K  \left(\exp \sum_t r_E(s^{\tau_k}_t, a^{\tau_k}_t)\right) / d_k
\end{equation*}  
where $d_k = \sum_{j=k}^K \exp \sum_{t} \gamma^t r_E(s^{\tau_j}_t, a^{\tau_j}_t)$. Then, we make the same substitution using the inverse bellman operator giving us the permutation model $P_Q$ implied by the Q function, and run maximum likelihood estimation over the model with the preference loss $\mathcal{L}_p (Q) = \mathbb{E}_{\tau \sim \mathcal{D}_p }\left[ \log P_Q (\tau) \right] + \lambda \psi(r)$.

\section{Results}
\label{app:results}
We divide this part of appendix into four different sections following the results section. Each section additionally provides hyper-parameters used for \abv in that section. The first section, setup, contains detailed information on the experimental setup and hyper-parameters used. The second section on benchmark results gives full learning curves for the experiments in Section 4.2. The third section provides full learning curves for the MetaWorld and Data-scaling experiments. The final Appendix section provides extended ablations.  

\subsection{Setup}
Here we provide the full algorithmic outline of \abv using Implicit Q-Learning \citep{kostrikov2021offline} that mimics our implementation. While in practice the policy $\pi$ could be extracted at the end of training, we do it simultaneously as in \citep{kostrikov2021offline} in order to construct learning curves.  
\begin{algorithm}[H]
\caption{IPL Algorithm (IQL Variant)}
\label{alg:ipl_iql}
\SetKwInOut{Input}{Input}
\Input{$\mathcal{D}_p$, $\mathcal{D}_o$, $\lambda$, $\alpha$}
\For{$i = 1, 2, 3, ...$}{
Sample batches $B_p \sim \mathcal{D}_p, B_o \sim \mathcal{D}_o$ \\
Update $Q$: $\min_Q \E_{B_p}[\mathcal{L}_p(Q)] + \lambda \E_{B_p \cup B_o}[\mathcal{L}_r(Q)]$ \\
Update $V$: $\min_V \E_{B_p \cup B_o}\left[\left|\tau - \vmathbb{1}(Q(s,a)-V(s))\right| \left(Q(s,a) - V(s)\right)^2\right]$ \\ 
Update $\pi$: $\max_\pi \E_{\mathcal{D}_p \cup \mathcal{D}_o}[e^{\beta(Q(s,a) - V(s))} \log \pi(a|s)]$
}
\end{algorithm}
Note that above we write the temperature parameter $\beta$ as done in IQL, instead of how it is usually done, using $\alpha$ in the denominator \citep{xql, peng2019advantage}.

When sampling batches of preference data $B_p \sim \mathcal{D}_p$, we take sub-samples of each segment $\sigma$ of length $s$. For a sampled data point $(\sigma^{(1)}, \sigma^{(2)}, y)$, we sample $\textrm{start} \sim \textrm{Unif}[0, 1, 2, ... k - s]$ and then let take $\sigma = s_{\textrm{start}}, a_{\textrm{start}}, ..., s_{\textrm{start} + s}$. We use the same start value across the entire batch.  

Given that we run experiments using MLPs, all of our experiments were run on CPU compute resources. Each seed for each method requires one CPU core and 8 Gb of memory.

\subsection{Benchmark Results}
Here we provide details for our experiments on the preference-based RL benchmark from \citet{kim2023preference}. We use the same hyperparameters as \citet{kim2023preference} and \citet{kostrikov2021offline} as shown in \cref{tab:benhcmark_params}. \looseness=-1

\begin{figure}[h]
\centering
\includegraphics[width=0.85\textwidth]{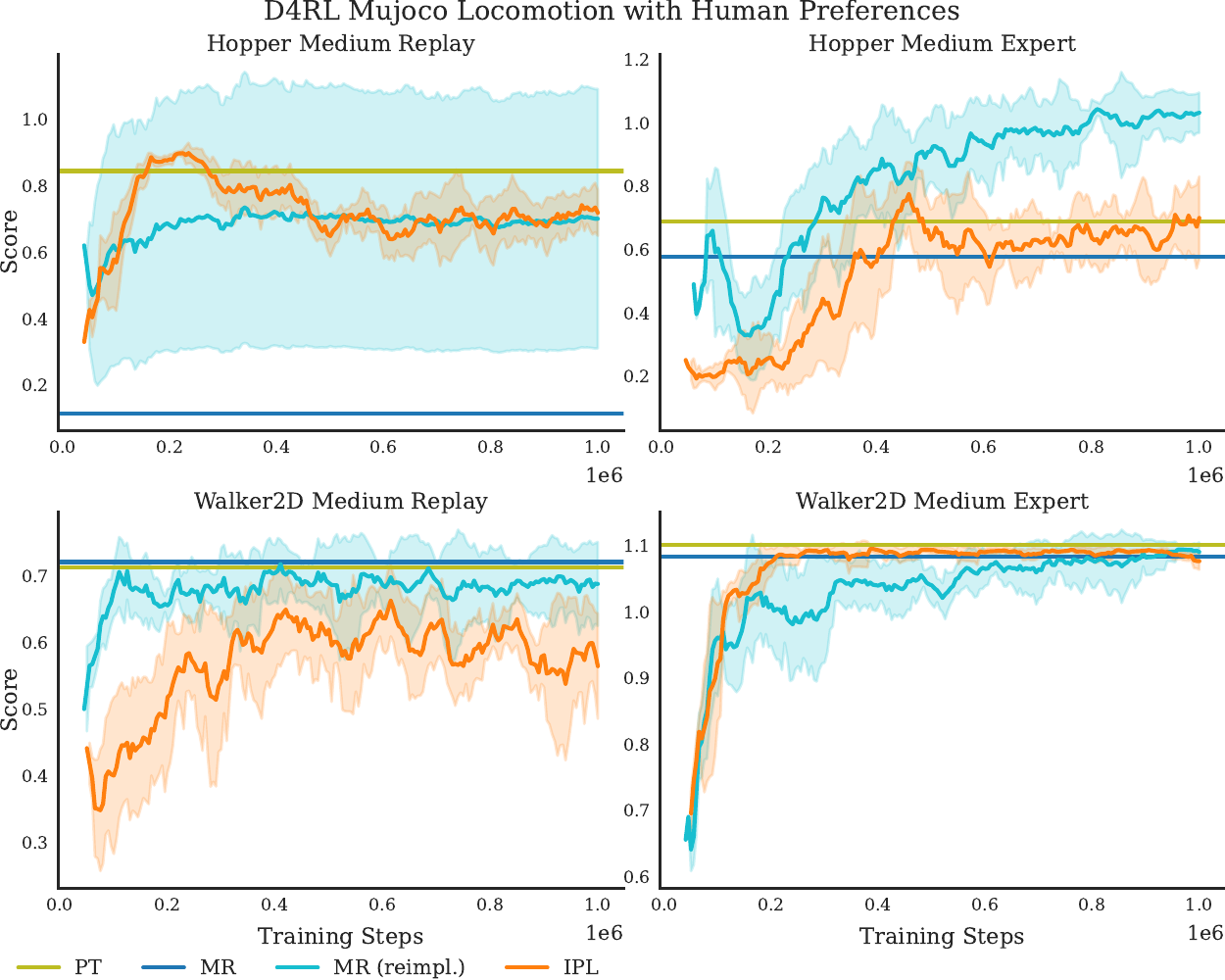}
\vspace{-0.05in}
\caption{Full learning curves on the D4RL locomotion benchmark with human preferences.}
\label{fig:locomotion}
\vspace{-0.1in}
\end{figure}

\textbf{Gym-Mujoco Locomotion}. Hopper and Walker2D agents are tasked with learning locomotion policies from datasets of varying qualities taken from the D4RL \citep{fu2020d4rl} benchmark. Preference datasets were constructed by \citet{kim2023preference} by uniformly sampling segments. Preference datasets for ``medium'' quality offline datasets  contain 500 queries, while preference datasets for ``expert'' quality offline datasets contain 100 queries. Segment length $k = 100$ for all datasets, and were subsampled to length $s = 64$ by \abv and our MR (reimpl). Evaluation was preformed over 10 episodes every 5000 steps. Full learning curves are shown in \cref{fig:locomotion}.

\begin{figure}[h]
\centering
\includegraphics[width=0.9\textwidth]{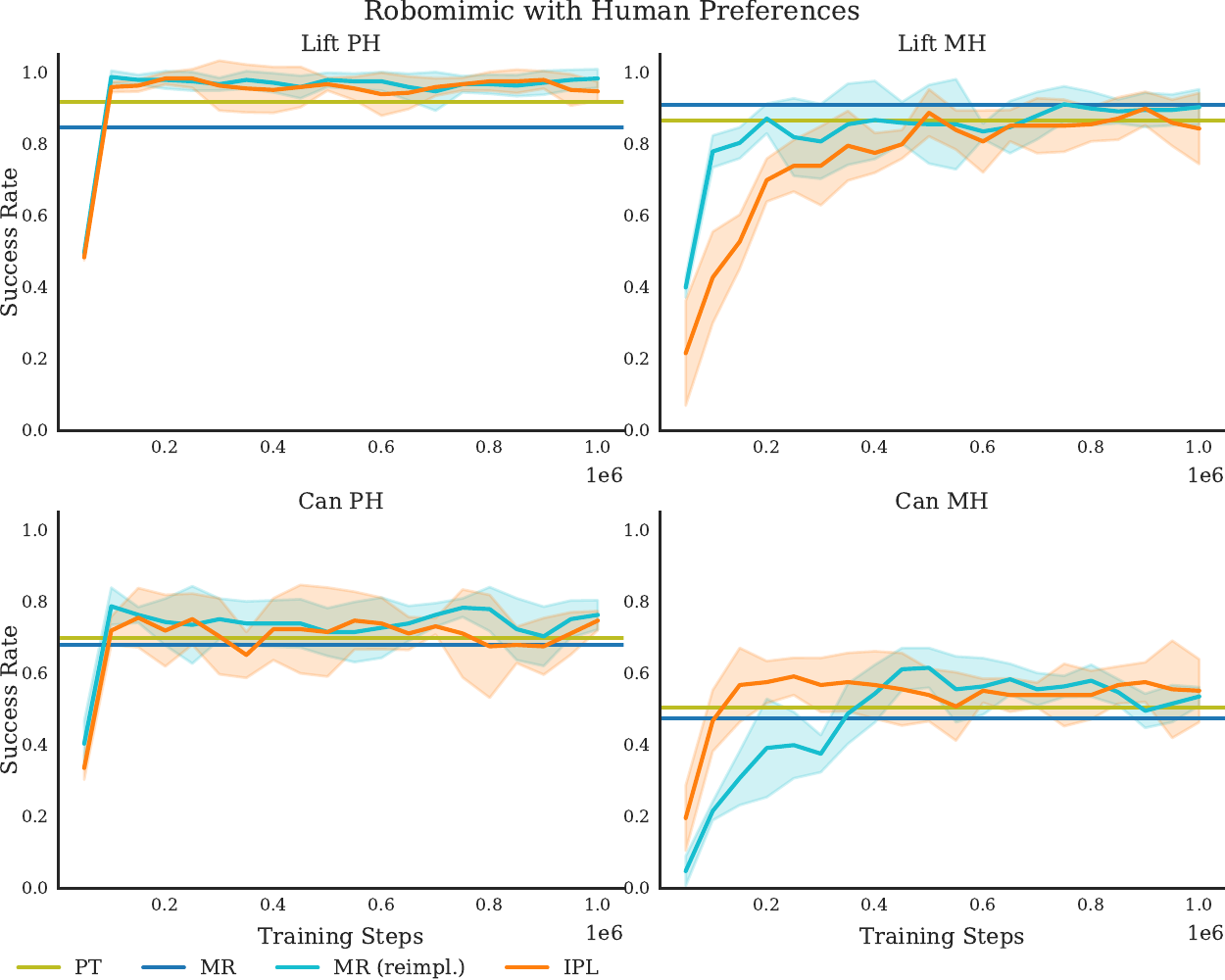}
\caption{Full learning curves on the RoboMimic benchmark with human preferences.}
\label{fig:robomimic}
\end{figure}

\textbf{RoboMimic}. The RoboMimic datasets contain interaction data of two types: ph --- proficient human and mh -- multihuman. The multi-human data was collected from human demonstrators of mixed quality. The robot is tasked with learning how to lift a cube (lift) or pick and place a can (can). Preference datasets were again taken directly from \citet{kim2023preference}. Preference datasets of size 100 with segment lengths $k = 50$, randomly sub-sampled to length $s = 32$ were used for the ph datasets. Preference datasets of size 500 with segment lengths $k = 100$, randomly sub-sampled to length $s = 64$ were used for the mh datasets. Evaluation was performed over 25 episodes every 50000 steps. Full learning curves are shown in \cref{fig:robomimic}.

\begin{figure}[h]
\centering
\includegraphics[width=0.75\textwidth]{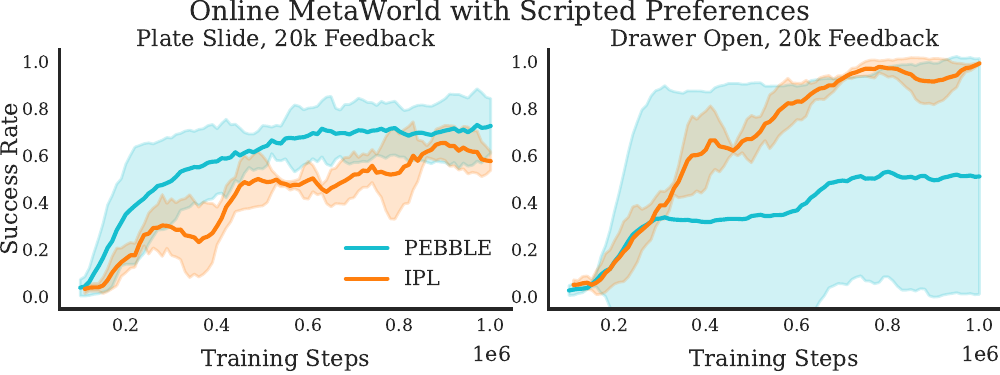}
\vspace{-0.1in}
\caption{Online experiments for two of the most difficult MetaWorld environments. Each method (run on 3 seeds) gets feedback every 5000 steps for a total of 20k artificially queries. Notice that IPL has much lower variance across runs, particularly in Drawer.}
\label{fig:online}
\vspace{-0.1in}
\end{figure}

\textbf{Online Experiments}. We also test a combination of \abv with PEBBLE \citep{pebble} on a few tasks in the MetaWorld benchmark. Results can be found in \cref{fig:online}

\begin{table}[H]
\centering
\begin{tabular}{lllllll}
\multicolumn{3}{c}{\textbf{Common Hyperparameters}}           &                      & \multicolumn{3}{c}{\textbf{MR Hyperparameters}}                 \\
\textbf{Parameter} & \textbf{Locomotion} & \textbf{Robomimic} &                      & \textbf{Parameter}   & \textbf{Locomotion} & \textbf{Robomimic} \\ \cline{1-3} \cline{5-7} 
$Q, V, \pi$ Arch   & 2x 256d             & 2x 256d            &                      & $r_\theta$ Arch      & 2x 256d             & 2x 256d            \\
Learning Rate      & 0.0003              & 0.0003             &                      & $r_\theta$ LR        & 0.0003              & 0.0003             \\
Optimizer          & Adam                & Adam               &                      & $r_\theta$ Optimizer & Adam                & Adam               \\
$\beta$            & 3.0                 & 0.5                &                      & $r_\theta$ Steps     & 20k                 & 20k                \\
$\tau$             & 0.7                 & 0.7                &                      &                      &                     &                    \\
$\mathcal{D}_o$ Batch Size & 256         & 256                &                      &                      &                     &                    \\
$\mathcal{D}_p$ Batch Size & 8           & 8                  &                      &                      &                     &                    \\
Training Steps     & 1 Mil               & 1 Mil              & \multicolumn{1}{c}{} & \multicolumn{3}{c}{\textbf{\abv Hyperparameters}}                \\
$k$                & 100                 & 100, 50            &                      & \textbf{Parameter}   & \textbf{Locomotion} & \textbf{Robomimic} \\ \cline{5-7} 
Subsample $s$      & 64                  & 64, 32             &                      & $\lambda$            & 0.5                 & 4                 
\end{tabular}
\vspace{0.05in}
\caption{Hyperparameters used for the benchmark experiments. We can see that \abv has fewer hyperparameters. For the $X$QL experiments we use $\alpha = 2$ for locomotion and $\alpha = 5$ for Robomimic. We left all other parameters the same.}
\label{tab:benhcmark_params}
\end{table}

\subsection{Data Scaling Results}
\begin{figure}[H]
\centering
\includegraphics[width=\textwidth]{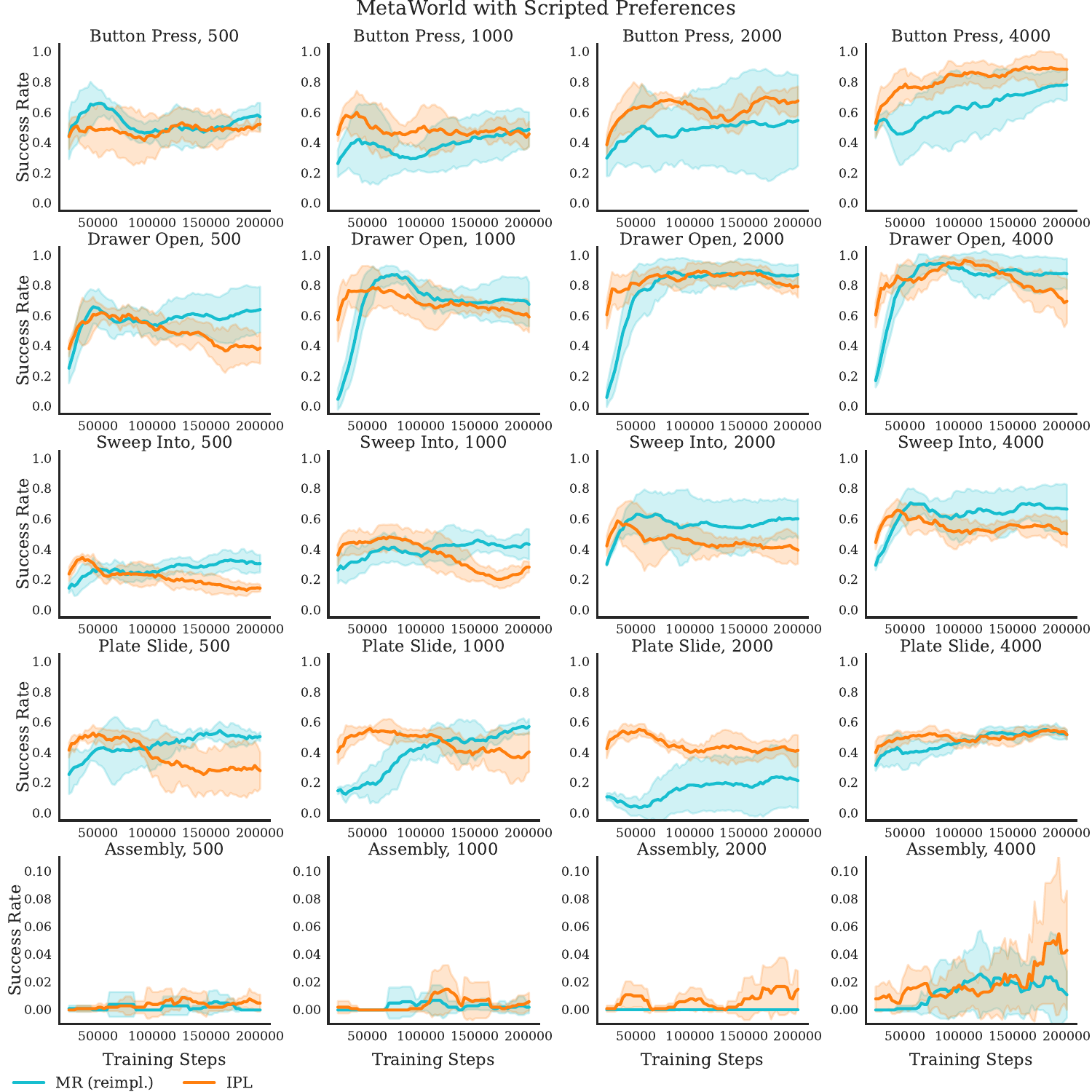}
\caption{Full learning curves for the MetaWorld data scaling results with scripted preferences.}
\label{fig:data_scaling}
\end{figure}

Experiments for data scaling were conducted on the MetaWorld benchmark from \citet{yu2020meta}. Offline datasets for five different MetaWorld tasks were constructed as follows: Collect 100 trajectories of expert data on the target task using the built in ground truth policies with the addition of Gaussian noise of standard deviation 1.0. Collect 100 trajectories of sub-optimal data by running the ground-truth policy for a different randomization of the target task with Gaussian noise 1.0. Collect 100 trajectories of even more sub-optimal data by running the ground truth policy \textit{of a different task} with Gaussian noise standard deviation 1.0 in the target domain. Finally, collect 100 trajectories with uniform random actions. As MetaWorld episodes are 500 steps long, this results in 200,000 time-steps of data. We then construct preference datasets by uniformly sampling segments from the offline dataset and assigning labels $y$ according to $\sum_t r(s_t^{(1)}, a_t^{(1)}) > \sum_t r(s_t^{(2)}, a_t^{(2)})$ where $r$ is the ground truth reward provided by metaworld. We then train using only the data from $\mathcal{D}$ General architecture hyper-parameters were taken from \citet{pebble, hejna2022fewshot} which also use the MetaWorld benchmark, but for online preference-based RL. Full-hyper parameters are shown in \cref{tab:data_params}. We run 20 evaluation episodes every 2500 steps. Full learning curves are shown in \cref{fig:data_scaling}. When reporting values in \cref{tab:data}, we choose the maximum point on the learning curves which average across five seeds. This provides results as if early stopping was given by an oracle, which is less optimistic than averaging the maximum of each seed as done in \citet{robomimic2021}.

\begin{table}[h]
\centering
\begin{tabular}{lllll}
\multicolumn{2}{c}{\textbf{Common Hyperparameters}} & \multicolumn{1}{c}{} & \multicolumn{2}{c}{\textbf{MR Hyperparameters}}  \\
\textbf{Parameter}         & \textbf{Value}         &                      & \textbf{Parameter}       & \textbf{Value}        \\ \cline{1-2} \cline{4-5} 
$Q, V, \pi$ Arch           & 3x 256d                &                      & $r_\theta$ Arch          & 3x 256d               \\
Learning Rate              & 0.0003                 &                      & $r_\theta$ LR            & 0.0003                \\
Optimizer                  & Adam                   &                      & $r_\theta$ Optimizer     & Adam                  \\
$\beta$                    & 4.0                    &                      & $r_\theta$ Steps         & 20k                   \\
$\tau$                     & 0.7                    &                      &                          &                       \\
$\mathcal{D}_p$ Batch Size & 16                     &                      &                          &                       \\
Training Steps             & 200k                   &                      & \multicolumn{2}{c}{\textbf{\abv Hyperparameters}} \\
$k$                        & 25                     &                      & \textbf{Parameter}       & \textbf{Value}   \\ \cline{4-5} 
Subsample $s$              & 16                     &                      & $\lambda$                & 0.5                  
\end{tabular}
\vspace{0.05in}
\caption{Hyper-parameters used in the MetaWorld data scaling experiments.}
\label{tab:data_params}
\end{table}

\subsection{Ablations}
In this section we provide additional ablations on both the benchmark datasets and MetaWorld datasets. We keep the hyperparameters the same, except for the parameter-efficient experiments. 

\begin{figure}[H]
\centering
\includegraphics[width=\textwidth]{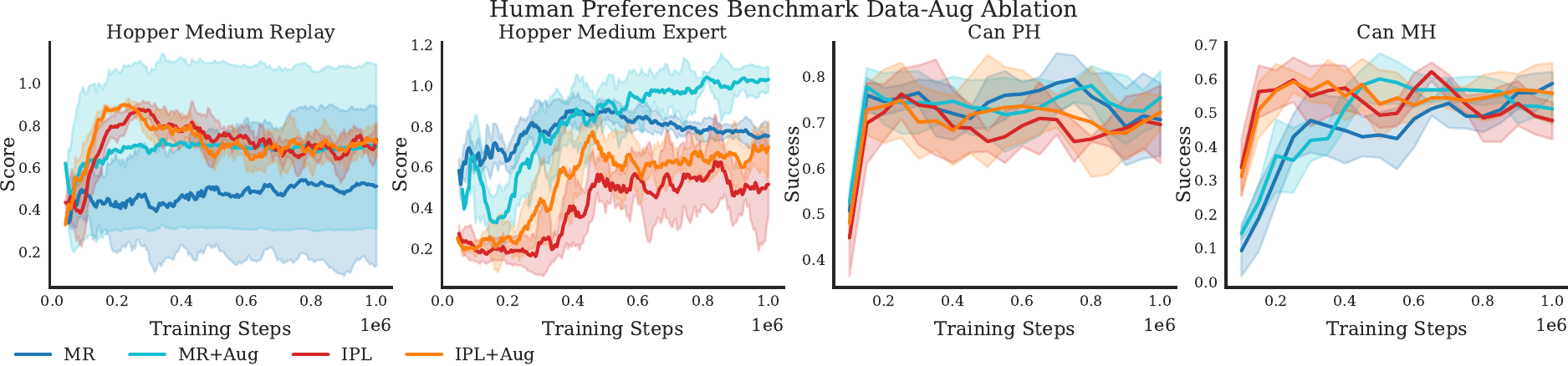}
\vspace{-0.1in}
\caption{IPL and MR+IQL with and without data augmentation across 5 seeds. We see that data augmentation makes a large difference, especially for MR+IQL in the hopper environment, while its effects are less for the robomimic Can datasets.}
\label{fig:augmentation}
\vspace{-0.1in}
\end{figure}

\textbf{Benchmark \abv Ablations.} We include results of full ablations for \abv on the benchmark tasks in \cref{tab:extended_ablations}. We additionally provide comparisons between \abv and MR + IQL with and without data augmentation in \cref{fig:augmentation}.

\begin{table}[H]
\centering
\begin{tabular}{lcccc}
\textbf{Dataset} & \textbf{No Aug}   & \textbf{$\lambda$ = 0} & \textbf{IPL-XQL}  & \textbf{IPL}      \\ \hline
hop-m-r          & 70.46 $\pm$ 6.73  & 10.41 $\pm$ 2.26       & 80.4 $\pm$ 2.13   & 73.57 $\pm$ 6.67  \\
hop-m-e          & 51.26 $\pm$ 17.46 & 52.81 $\pm$ 7.45       & 54.3 $\pm$ 12.33  & 74.52 $\pm$ 10.11 \\
walk-m-r         & 58.50 $\pm$ 5.31  & 4.85 $\pm$ 1.52        & 57.82 $\pm$ 5.24  & 59.92 $\pm$ 5.11  \\
walk-m-e         & 108.91 $\pm$ 0.18 & 58.77 $\pm$ 15.75      & 75.16 $\pm$ 23.40 & 108.51 $\pm$ 0.60 \\
lift-ph          & 98.0 $\pm$ 2.53   & 85.2 $\pm$ 7.71        & 98.40 $\pm$ 2.59  & 97.60 $\pm$ 2.94  \\
lift-mh          & 84.8 $\pm$ 4.11   & 52.60 $\pm$ 10.07      & 89.00 $\pm$ 4.37  & 87.20 $\pm$ 5.31  \\
can-ph           & 68.6 $\pm$ 8.25   & 25.4 $\pm$ 5.25        & 68.6 $\pm$ 7.66   & 74.8 $\pm$ 2.40   \\
can-mh           & 53.2 $\pm$ 5.8    & 13.8 $\pm$ 5.73        & 59.0 $\pm$ 5.0    & 57.6 $\pm$ 5.00  
\end{tabular}
\caption{Extended \abv ablation results.}
\label{tab:extended_ablations}
\end{table}

\begin{figure}[h]
\centering
\includegraphics[width=\textwidth]{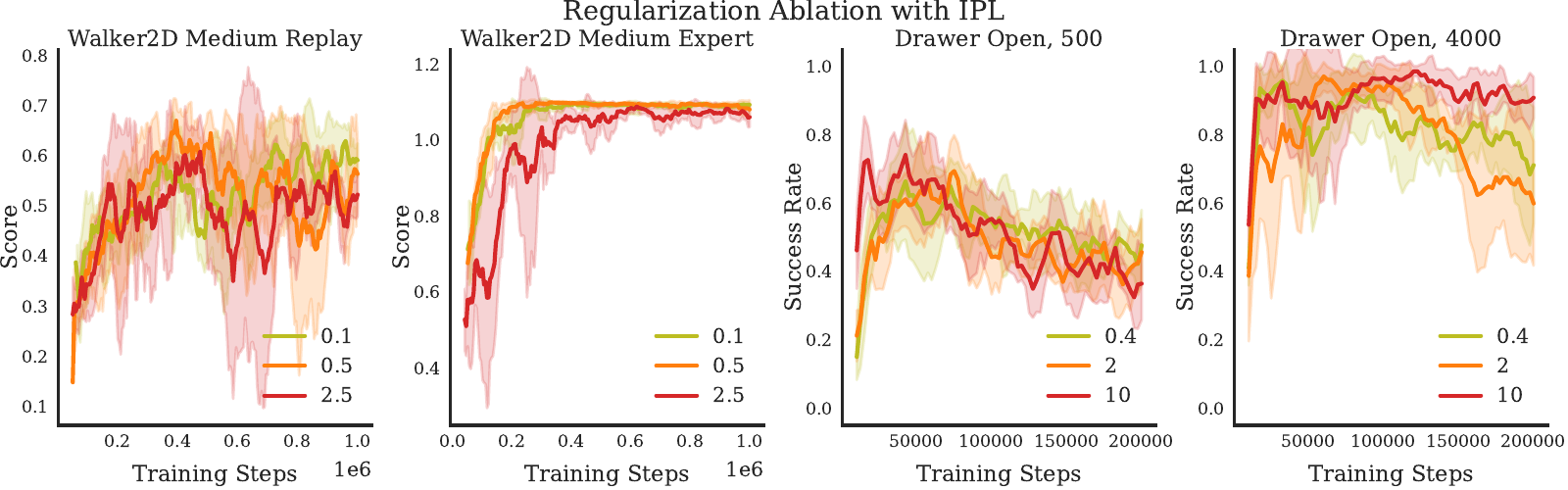}
\vspace{0.05in}
\includegraphics[width=\textwidth]{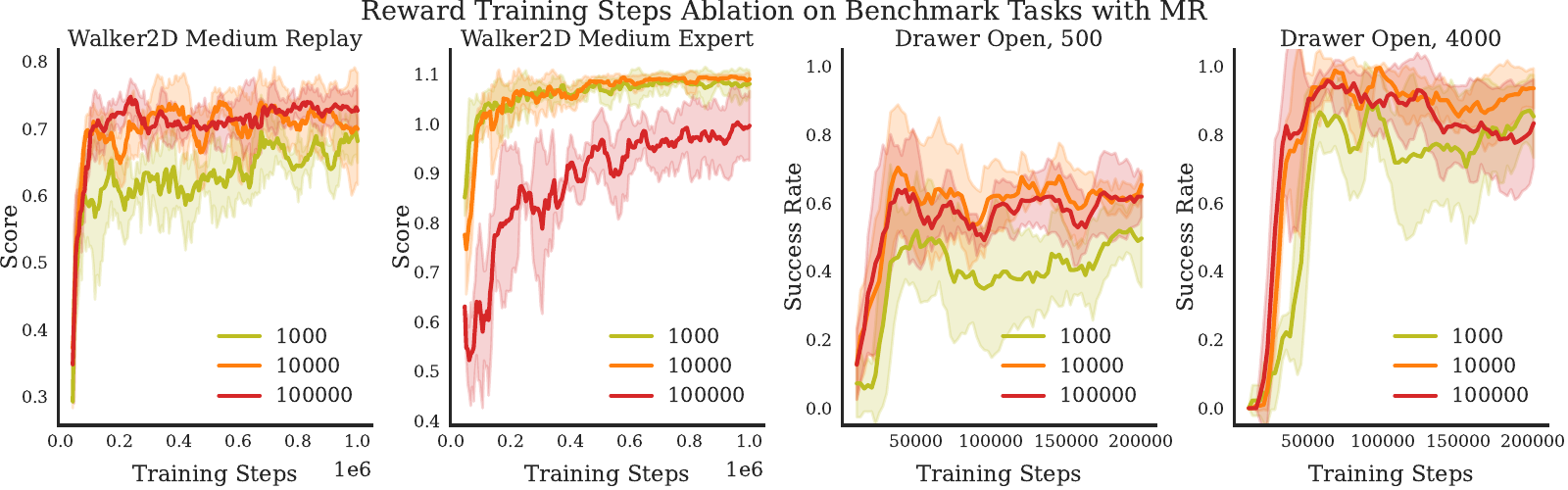}
\caption{Ablations on regularization strength $\lambda$ for \abv (top row) and the number of reward steps for MR (bottom row). We see that \abv is relatively consistent across different values of $\lambda$. MR on the other hand, can vary greatly if the reward function under or over fits. In Walker2D Medium Replay and Drawer Open, 500, we see that it can easily under-fit. In Walker2D Medium Expert it easily over-fits.}
\label{fig:full_ablations}
\end{figure}

\textbf{Hyper-parameter Sensitivity.}
We run hyper-parameter sensitivty results for the human-preference benchmark datasets in \cref{fig:full_ablations}. The top row depicts the sensitivity for \abv to the value of $\lambda$. The bottom row depicts the sensitivity of MR to the number of timesteps the reward function is trained for. 

\textbf{Parameter Efficiency.}
For the parameter-efficient experiments \textit{only} we use an efficient version of \abv based on AWAC \citep{nair2021awac} to additionally remove the need for learning value network. AWAC uses a policy-evaluation, policy imporvement style approach. The outline of this variant is given below:
\begin{algorithm}[H]
\caption{IPL Algorithm (AWAC Variant)}
\label{alg:ipl_awac}
\SetKwInOut{Input}{Input}
\Input{$\mathcal{D}_p$, $\mathcal{D}_o$, $\lambda$, $\alpha$}
\For{$i = 1, 2, 3, ...$}{
Sample batches $B_p \sim \mathcal{D}_p, B_o \sim \mathcal{D}_o$ \\
Estimate $V$ as $Q(s, \pi(s))$ \\
Update $Q$: $\min_Q \E_{B_p}[\mathcal{L}_p(Q)] + \lambda \E_{B_p \cup B_o}[\mathcal{L}_r(Q)]$ \\
Update $\pi$: $\max_\pi \E_{\mathcal{D}_p \cup \mathcal{D}_o}[e^{\beta(Q(s,a) - Q(s, \pi(s)))} \log \pi(a|s)]$
}
\end{algorithm}
For this version of \abv, we use $\lambda = 0.5$. All other hyper-parameters remain the same as in Table \ref{tab:parameter_efficiency} except the architectures. For the parameter-efficiency experiments only we use MLPs consisting of two dense layers with either dimension 64 or dimension 35. Running MR with a two-layer MLP of dimension 35 has almost exactly the same number of parameters as \abv-AWAC with two-layer MLPs of dimension 64. We include full results for the parameter-efficiency experiments in \cref{tab:parameter_efficiency}. We find that on Drawer Open and Sweep Into, \abv outperforms both MR (64) and MR (35). In these environments, performance increases from MR (35) to MR (64) indicating that the expressiveness of the $Q$-function and policy are limiting performance. For the same budget, \abv is able to perform better. In Button Press, the simplest task, we find that MR (64) actually over-fits more than MR (35) and MR (64) ends up performing worse. In Plate Slide, all methods perform similarly independent of parameter count. We omit Assembly because of its low success rate at all data scales.

\begin{table}[h]
\centering
\begin{tabular}{llcccc}
\multicolumn{2}{r}{Preference Queries}   & \multicolumn{1}{r}{500}  & \multicolumn{1}{r}{1000} & \multicolumn{1}{r}{2000} & \multicolumn{1}{r}{4000} \\ [0.1cm] \hline
\multirow{3}{*}{Button Press} & MR (35)  & \textbf{73.9 \stdv{8.9}}  & \textbf{86.8 \stdv{8.2}}  & \textbf{89.9 \stdv{14.4}} & \textbf{99.0 \stdv{1.0}}  \\
                              & MR (64)  & 54.2 \stdv{16.1}          & 42.6 \stdv{33.0}          & 67.1 \stdv{14.9}          & 43.4 \stdv{7.4}           \\
                              & IPL (64) & 65.8 \stdv{13.3}          & 79.8 \stdv{18.1}          & 80.0 \stdv{17.3}          & \textbf{95.8 \stdv{5.2}}  \\ [0.1cm]
\multirow{3}{*}{Drawer Open}  & MR (35)  & 13.4 \stdv{13.9}          & 12.6 \stdv{21.9}          & 15.5 \stdv{20.1}          & 18.4 \stdv{25.6}          \\
                              & MR (64)  & 13.4 \stdv{19.0}          & 57.1 \stdv{31.2}          & 54.5 \stdv{31.7}          & 78.8 \stdv{12.2}          \\
                              & IPL (64) & \textbf{89.8 \stdv{11.3}} & \textbf{93.2 \stdv{2.5}}  & \textbf{99.5 \stdv{0.9}}  & \textbf{95.5 \stdv{3.7}}  \\ [0.1cm]
\multirow{3}{*}{Sweep Into}   & MR (35)  & 35.1 \stdv{8.9}           & 42.4 \stdv{9.9}           & 45.9 \stdv{9.6}           & 35.9 \stdv{4.1}           \\
                              & MR (64)  & 31.1 \stdv{6.4}           & 55.8 \stdv{5.9}           & 49.6 \stdv{10.3}          & 56.4 \stdv{10.3}          \\
                              & IPL (64) & \textbf{41.1 \stdv{14.2}} & \textbf{63.9 \stdv{8.0}}  & \textbf{65.0 \stdv{12.0}} & \textbf{63.9 \stdv{11.8}} \\ [0.1cm]
\multirow{3}{*}{Plate Slide}  & MR (35)  & \textbf{55.2 \stdv{6.1}}  & \textbf{51.1 \stdv{4.4}}  & \textbf{53.0 \stdv{2.0}}  & \textbf{48.9 \stdv{3.3}}  \\
                              & MR (64)  & 46.6 \stdv{21.9}          & \textbf{50.8 \stdv{0.6}}  & 47.0 \stdv{2.5}           & \textbf{48.5 \stdv{4.6}}  \\
                              & IPL (64) & \textbf{54.9 \stdv{3.2}}  & \textbf{49.4 \stdv{1.6}}  & 45.2 \stdv{9.0}           & \textbf{48.8 \stdv{4.9}} 
\end{tabular}
\caption{Performance of different methods on the MetaWorld tasks under a limited parameter budget. MR (35) and IPL (64) have the same number of parameters. The Assembly task is ommited due to low success rate.
On Button Press, fewer parameters appears to perform better as, due to the simplicity of the task, its easier for the bigger models to overfit. On Drawer Open and Sweep Into, we see consistent gains from increasing
the number of parameters in the network, and \abv performs best overall. On the Plate Slide task, all methods at different parameter scales perform similarly. }
\label{tab:parameter_efficiency}
\end{table}

\end{document}